\documentclass[iicol,sn-basic]{sn-jnl}

\usepackage{graphicx}
\usepackage{amsmath}
\usepackage{amssymb}
\usepackage{csquotes}
\usepackage{url}
\usepackage{booktabs}
\usepackage[table]{xcolor}
\usepackage{colortbl}
\usepackage{algorithm}
\usepackage{algorithmicx}
\usepackage{algpseudocode}
\usepackage[nolist]{acronym}

\usepackage[inline]{enumitem}
\usepackage{multirow}
\usepackage[separate-uncertainty=true]{siunitx}

\algnewcommand\algorithmicforeach{\textbf{for each}}
\algdef{S}[FOR]{ForEach}[1]{\algorithmicforeach\ #1\ \algorithmicdo}

\newacro{dnn}[DNN]{Deep Neural Network}
\newacro{fcn}[FCN]{Fully Convolutional Network}
\newacro{pc}[PC]{Point Cloud}
\newacro{sdf}[SDF]{Signed Distance Function}
\newacro{cnn}[CNN]{Convolutional Neural Network}
\newacro{gnn}[GNN]{Graph Neural Network}
\newacro{dl}[DL]{Deep Learning}
\newacro{ml}[ML]{Machine Learning}
\newacro{gpis}[GPIS]{Gaussian Process Implicit Surface}
\newacro{mc}[MC]{Monte Carlo}
\newacro{mlp}[MLP]{Multi-Layer Perceptron}
\newacro{bpa}[BPA]{Ball Pivoting Algorithm}
\acrodefplural{gpis}[GPIS's]{Gaussian Process Implicit Surfaces}
\newacro{gpisp}[GPISP]{Gaussian Process Implicit Shape Potential}
\newacro{gp}[GP]{Gaussian Process}
\acrodefplural{gp}[GPs]{Gaussian Processes}
\newacro{ros}[ROS]{Robot Operating System}
\newacro{icp}[ICP]{Iterative Closest Point}
\newacro{fps}[FPS]{Farthest Point Sampling}
\newacro{igr}[IGR]{Implicit Geometric Regularization for Learning Shapes}
\newacro{js}[JS]{Jaccard similarity}
\newacro{cd}[CD]{Chamfer distance}
\newacro{hd}[HD]{Hausdorff distance}

\newacro{nerf}[NeRF]{Neural Radiance Fields}
\newacro{dof}[DoF]{Degree of Freedom}
\newacro{sd}[SD]{Standard deviation}
\newacro{gsr}[GSR]{Grasp Success Rate}
\newacro{vqdif}[VQDIF]{Vector Quantized Deep Implicit Functions}

\newacro{tp}[TP]{True Positive}
\newacro{fp}[FP]{False Positive}
\newacro{fn}[FN]{False Negative}

\newcommand{\equationref}[1]{\hyperref[#1]{Eq.~\ref*{#1}}}
\newcommand{\figref}[1]{\hyperref[#1]{Fig.~\ref*{#1}}}
\newcommand{\tabref}[1]{\hyperref[#1]{Table~\ref*{#1}}}
\newcommand{\secref}[1]{\hyperref[#1]{Section~\ref*{#1}}}
\newcommand{\algoref}[1]{\hyperref[#1]{Alg.~\ref*{#1}}}

\newcommand{\norm}[1]{\left\lVert#1\right\rVert}

\DeclareMathOperator{\E}{\mathbb{E}}
\newcommand{\abs}[1]{\left\lvert#1\right\rvert}

\newcommand{\etal}[1]{#1 et al.}

\def\methodname{\textit{ShapeGrasp}} 
\def\methodnameo{\methodname$_1$} 
\def\methodnamen{\methodname} 

\def\sota{state-of-the-art}
\def\sotan{state of the art} 

\def\graspit{GraspIt!}

\def\gt{ground truth}

 \usepackage[firstpageonly=true]{draftwatermark}

\SetWatermarkAngle{0}
\SetWatermarkColor{black}
\SetWatermarkLightness{0.5}
\SetWatermarkFontSize{9pt}
\SetWatermarkVerCenter{30pt}
\SetWatermarkText{\parbox{30cm}{%
\centering This is the author's manuscript submitted for peer review.
\centering
}}

\title[ShapeGrasp]{ShapeGrasp: Simultaneous Visuo-Haptic Shape Completion and Grasping for Improved Robot Manipulation}

\author*[1]{\fnm{Lukas} \sur{Rustler}}\email{lukas.rustler@fel.cvut.cz}
\author*[1]{\fnm{Matej} \sur{Hoffmann}}\email{matej.hoffmann@fel.cvut.cz}

\affil[1]{\orgdiv{Department of Cybernetics, Faculty of Electrical Engineering}, \orgname{Czech Technical University in Prague}, \orgaddress{\city{Prague}, \country{Czech Republic}}}

\abstract{Humans grasp unfamiliar objects by combining an initial visual estimate with tactile and proprioceptive feedback during interaction. We present ShapeGrasp, a robotic implementation of this approach. The proposed method is an iterative grasp-and-complete pipeline that couples implicit surface visuo-haptic shape completion (creation of full 3D shape from partial information) with physics-based grasp planning. From a single RGB-D view, ShapeGrasp infers a complete shape (point cloud or triangular mesh), generates candidate grasps via rigid-body simulation, and executes the best feasible grasp. Each grasp attempt yields additional geometric constraints---tactile surface contacts and space occupied by the gripper body---which are fused to update the object shape. Failures trigger pose re-estimation and regrasping using the refined shape. We evaluate ShapeGrasp in the real world using two different robots and grippers. To the best of our knowledge, this is the first approach that updates shape representations following a real-world grasp. We achieved superior results over baselines for both grippers (grasp success rate of 84\% with a three-finger gripper and 91\% with a two-finger gripper), while improving the 3D shape reconstruction quality in all evaluation metrics used.}


\begin{document}

\maketitle

\section{Introduction}
Robust robot grasping of a priori unknown objects remains challenging whenever the object is only partially observed. A single RGB-D view typically yields a partial point cloud with large occluded regions. Shape completion methods can infer the missing geometry and provide full 3D shape, but the resulting model is uncertain, and grasp planning can fail when hallucinated geometry is incorrect or when the object is compliant.

At the same time, grasping is an informative interaction. During approach and closure, the gripper geometry constrains which regions of space must be empty in the object-centric representation (we call this free space), and after contact, the tactile sensors provide evidence of the true surface. If an initial grasp fails, these constraints can be exploited immediately to update the object model and re-plan the grasp, rather than repeatedly attempting grasps from the same visual-only observation.

\begin{figure*}[htb]
    \centering
    \includegraphics[width=1\textwidth]{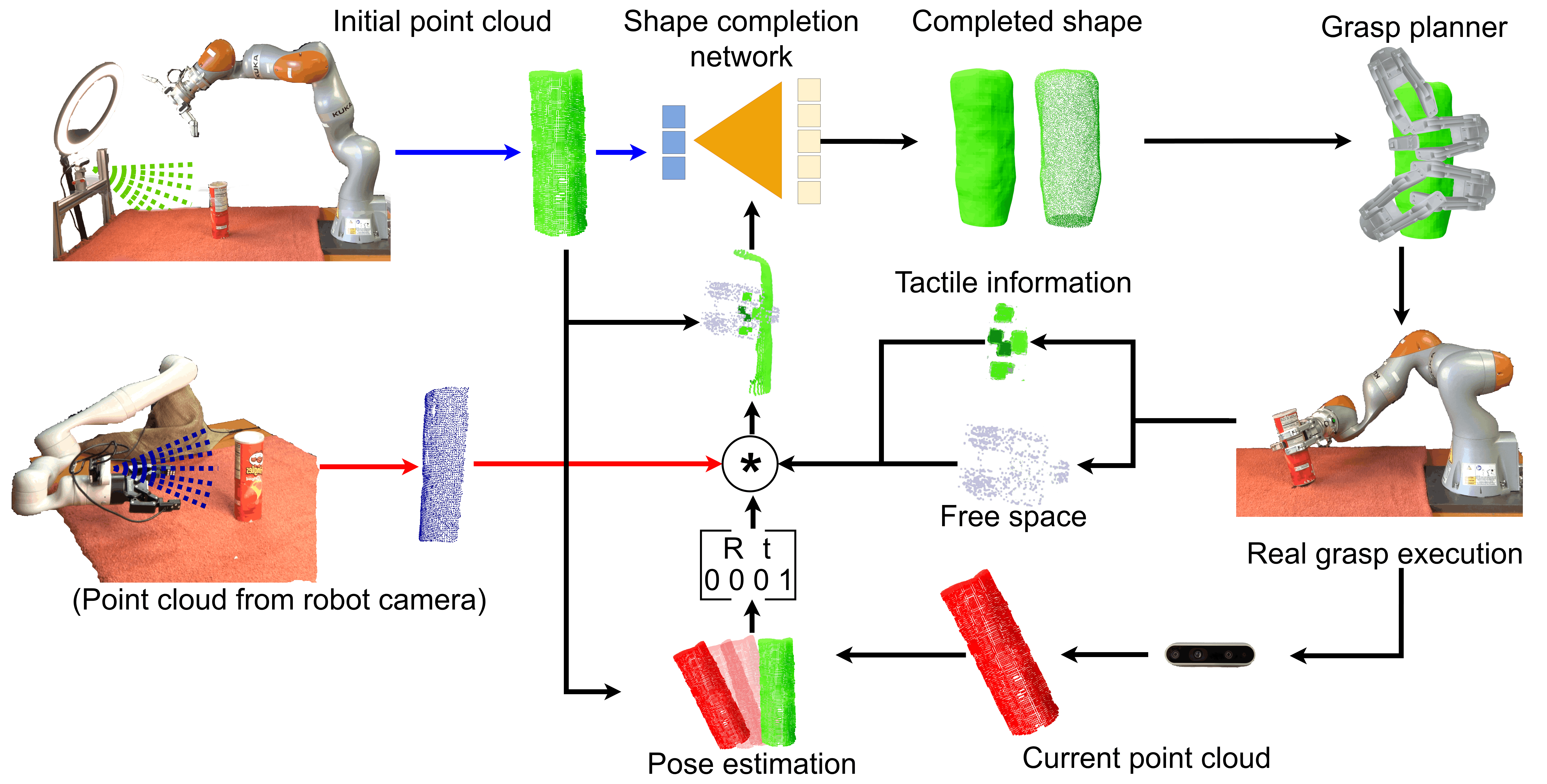}
    \caption{Schematic operation of \methodname{}. RGB-D information from a fixed camera is used to create a initial point cloud $\mathcal{X}$ (blue line; done only once) and later to track the pose of the object. $\mathcal{X}$ is inserted into a deep neural network (blue line; done only once) to compute a completed shape $O_0$. The shape is inserted into the proposed grasp planner and the best feasible grasp is executed. After the grasp, tactile information $\textbf{t}_r$ and free space information $\textbf{fs}_r$, where $r$ is the iteration number, is extracted from the gripper, transformed using the current pose of the object $\textbf{T}_r$ and inserted in $\mathcal{X}$. Optionally (red lines), point cloud from a wrist camera can be extracted, transformed with $\textbf{T}_r$ and also inserted to $\mathcal{X}$. New shape $O_r$ is computed. If the grasp was successful, the pipeline is ended. Otherwise, the new grasps are generated using the updated model $O_r$. The pipeline runs until successful grasp or until the maximum number of iterations $R$ is reached.}
    \label{fig:schema}
\end{figure*}

We study the following scenario. An unknown object is placed on a table in front of a robot arm. The system receives an initial RGB-D observation from a fixed camera and should grasp the object and simultaneously reconstruct a detailed 3D model that can be used for downstream estimation (e.g., volume, center of mass, or contact planning). Since a grasp can fail and move the object, the pipeline must also (i) detect failures, (ii) re-estimate the object pose, and (iii) regrasp using the updated shape. The described problem can be represented in the real world as a case where objects are presented to a system that is required to grasp them and, for example, sort them by some of their properties. Imagine a waste sorting factory with a conveyor belt that sort bottles into two boxes, with and without liquid inside. Other scenario can be robot-human handover, where robot needs to grasp unknown object and simultaneously needs to know its weight and center of mass for precise handover.

We draw inspiration from human manipulation. \citet{johansson2009coding} describe control that proceeds through phases (reach, grasp, lift, \ldots) with predictions of sensory outcomes. When sensory feedback does not match the prediction, humans adapt the internal model instead of blindly continuing to the next phase. Furthermore, our method is consistent with the idea of Active Perception (AP), as defined by Ruzena Bajcsy~\citeyearpar{Bajcsy88Active}, as we deliberately interact with and explore the environment to lower the uncertainty about the perceived object---by grasping the object to expose more of its surface and improve model accuracy.

We present \methodname{}, an iterative visuo-haptic pipeline that couples implicit-surface shape completion with physics-based grasp planning. Starting from the initial partial point cloud, we infer a watertight mesh and generate grasp candidates via a rigid-body simulation. After each grasp attempt, we add: (a) tactile surface points when sensors are activated, and (b) free space constraints derived from the space occupied by the gripper body. The object pose is tracked between iterations using RGB segmentation and point-cloud alignment, allowing newly acquired data to be fused in a common frame.

\textbf{Contributions.} The main contributions are:
\begin{enumerate}[label=(\roman*)]
    \item an iterative \emph{grasp-and-complete} pipeline that updates an implicit object model using tactile contacts and gripper-derived free space constraints collected during grasp attempts, and triggers regrasping when failures are detected. To the best of our knowledge, we are the first to improve the shape after grasping in the real world;
    \item a physics-based grasp proposal generator that consumes a reconstructed triangular mesh and supports multi-finger grippers via URDF models, enabling tight integration with the shape completion output;
    \item a real-world evaluation using novel objects on two robots (Kinova Gen3 with a Robotiq 2F-85 gripper with Contactile PapillArray sensor and KUKA iiwa with a Barrett Hand with capacitive tactile arrays) demonstrating improvements from \sota{} approaches in completion metrics and grasp success, together with an ablation study on the information sources used for completion;
    \item data and code are publicly available at \url{https://rustlluk.github.io/ShapeGrasp}.
\end{enumerate}

 \section{Related work}

\subsection{Shape Completion}
Shape completion aims to infer a complete object representation from partial observations. Usually, complete shapes are represented as point clouds, voxel grids, or triangular meshes. The completion approaches can be divided into three sections based on the modality used at input:
\begin{enumerate*}[label=(\roman*)]
    \item visual;
    \item haptic;
    \item visuo-haptic.
\end{enumerate*}

\textbf{Visual only.} The first approaches were visual-only. The early approaches were template-based and relied on a database of objects~\citep{pauly2005example}, utilized primitive shapes~\citep{schnabel2009completion}, or assumed the symmetry of all objects used~\citep{bohg2011mind}. As expected, those worked well for known objects but performed worse on new, unknown ones. Later, with advances in \ac{ml}, new methods based on \ac{dl} were proposed. Voxel grids allow for a more powerful generalization~\citep{varleyShapeCompletionEnabled2017,lundellRobustGraspPlanning2019}. However, the inclusion of voxel grids results in lower resolution due to memory and performance requirements. More recent approaches use latent space representation of objects~\citep{groppImplicitGeometricRegularization2020,parkDeepSDFLearningContinuous2019,atzmonSALSignAgnostic2020}, Graph Neural Networks~\citep{huang2021gascn} or Transformers~\citep{Rosasco2022, yan2022ShapeFormerTransformerBasedShapea, mohammadi_3dsgrasp_2023}.

\textbf{Tactile only.} Getting sufficient information only from vision is not always possible (e.g., in dark environments or with adversarial objects), and other modalities, such as touch, can bring an advantage. Touch information is mostly local---small area is explored---and thus purely haptic completion requires a high number of contacts and is not as common. Classical approaches include implicit shape potentials~\citep{ottenhausLocalImplicitSurface2016} and \ac{gp}~\citep{yiActiveTactileObject2016, driessActiveLearningQuery2017, dragiev2013uncertainty}. Gaussian methods provide mathematically solid solutions and also naturally include uncertainty as each point is modeled with the Gaussian distribution. The disadvantage is the need for a high number of touches/exploration actions. \etal{Bonzini} proposed to solve this using symmetry~\citeyearpar{Bonzini2022} and with a novel formulation of \ac{gp} models for modeling symmetric surfaces~\citeyearpar{2025Bonzini_autonomous_symmetry}.

\textbf{Visuo-Tactile.} The combination of the two modalities is arguably the most promising approach. Similarly to tactile-only approaches, \ac{gp} methods were proposed~\citep{gandlerObjectShapeEstimation2020,ottenhausVisuoHapticGraspingUnknown2019,bjorkmanEnhancingVisualPerception2013, Suresh2022}. Even with visual information, the methods still require a substantial number of touches. \ac{cnn}-based methods were proposed to solve this problem~\citep{watkins-vallsMultiModalGeometricLearning2019,wang3DShapePerception2018a}. These require fewer touches, but suffer from lower resolution due to computational requirements. Methods with higher resolution were proposed by \citet{smith3DShapeReconstruction2020,smithActive3DShape2021}, \citet{Murali2022} or \citet{Rustler2022ActVH, Rustler2023VISHAC}.

\subsection{Grasping}
Classical grasp analysis evaluates stability via geometric-
and force-closure~\citep{bicchi1995ClosurePropertiesRobotic, dizioglu1984MechanicsFormClosure, li2003ComputingThreefingerForceclosurea}, with widely used quality metrics
such as antipodality~\citep{nguyen1987ConstructingStableGrasps} and wrench-space measures~\citep{ferrari1992PlanningOptimalGrasps}. Wrench-space metrics require computing the convex hull of wrenches applied at each contact point. It is complex mainly in the case of 3D objects, as the wrenches---and hulls--are 6D. One of the arguably most well known grasping pipelines is the \graspit{} simulator~\citep{miller2004GraspIt} from Columbia University that also utilizes these metrics for grasp assessment. These metrics and methods require triangular meshes as input.

As in almost every field of robotics, grasping was also influenced by learning. One of the first \ac{dl} approaches was published by \citet{lenz2015DeepLearningDetecting}. They proposed to learn the \enquote{rectangle representation} using a deep network. A notable DL method by ten \citet{tenpas2017GraspPoseDetection} evaluates grasp poses by projecting local point cloud geometry into 2D multi-channel images and processing them with a CNN. The benefits of this approach are fast generation times and the ability to work effectively with single-view point clouds. Building on this pipeline, \citet{liang2019PointNetGPDDetectingGrasp} proposed PointNetGPD, an approach that evaluates the raw 3D point clouds directly. This architecture requires significantly fewer parameters than traditional CNNs and outputs a continuous probability, or quality score, for the generated grasps. A well-known family of approaches \textit{Dex-Net} was presented by \etal{Mahler}, starting with~\citeyearpar{mahler2016DexNetCloudbasedNetwork}. The most recent approaches use, for example, diffusion models~\citep{weng_dexdiffuser_2024} or \acp{gnn}~\citep{huang2023DefGraspNetsGraspPlanning}. Most of these expect a point cloud on the input; either single- or multi-view.

Today, the field is taking advantage of advances in computational power and simulation. \etal{Sundermeyer} was able to train 6-\ac{dof} grasp generation deep network on 17 million simulated grasps~\citeyearpar{sundermeyer2021ContactGraspNetEfficient6DoF}. Other researchers are using GPU-accelerated simulation to simulate the outcomes of grasping deformable objects~\citep{huang2022DefGraspSimPhysicsBasedSimulation, le2022NovelSimulationBasedQuality, huang2023DefGraspNetsGraspPlanning}. Interestingly, using simulation is making 3D triangular meshes valuable again.

Our proposed grasp planner is on the boundary of the old-fashion and new wave. We do not utilize deformable objects and GPU-parallelization and instead stick with simpler (and less demanding) CPU-computed physics. On the other hand, we utilize the advantages of triangular meshes that allow for better grasp evaluation.

\subsection{Grasping Using Shape Completion}
Grasping frequently serves as a primary objective—or at least a critical evaluation metric—for 3D shape completion. Highlighting this connection, \citet{varleyShapeCompletionEnabled2017} used \graspit{} to demonstrate that completing shapes increases the grasp success rates by approximately 20\%. Building on this foundation, \citet{lundellRobustGraspPlanning2019} benchmarked their completion network against Varley's using \graspit{}, a simulator that was similarly employed to validate shape completions by \citet{Rustler2022ActVH, Rustler2023VISHAC}. Transitioning to other grasp planners, \citet{Rosasco2022} evaluated their method against Lundell's by leveraging grasp candidates generated by GPD~\citep{tenpas2017GraspPoseDetection}. \citet{mohammadi_3dsgrasp_2023} also integrated their completion network with GPD, an approach that \citet{duarte_measuring_2025} subsequently improved by incorporating uncertainty into the grasp scoring pipeline.

Finally, several works have explored tightly integrated or highly efficient architectures for simultaneous completion and grasping. \citet{van_der_merwe_learning_2020} introduced \textit{PointSDF} to learn \ac{sdf} embeddings for partial point clouds, utilizing the same architecture for geometry-aware grasp success prediction. In the work of \citet{yang_robotic_2021}, grasp predictions from a ResNet-34 model~\citep{he2016ResNet} are refined by projecting them directly onto shapes completed via the work of ~\citet{mitchell2020higher}.
More recently, the pipeline proposed by \citet{humt_combining_2023} achieves high efficiency by combining \ac{vqdif}-based completions~\citep{yan2022ShapeFormerTransformerBasedShapea} with a multi-finger grasp regressor, successfully generating both the complete shape and the grasp pose in under \qty{1}{\second}. \citet{de_farias_simultaneous_2021} proposed a method similar to ours that, in a simulation, reconstructs the initial shape using \ac{gpis} and then optimizes the placement of the fingers to iteratively obtain a higher probability of force-closure while collecting tactile data to improve the complete shape. The same main authors further proposed a task-informed grasping pipeline also based on initial shape completion using \ac{gpis}~\citep{de_farias_task-informed_2024}.

Our work combines the shape completion and grasping part together providing a symbiotic pipeline. \section{Method}
We propose an (iterative) pipeline that integrates visuo-haptic shape completion and grasping---depicted in \figref{fig:schema}. The objective is to grasp an arbitrary object on a table in front of a robotic manipulator while simultaneously completing the shape of the object. The entire pipeline is described in \secref{sec:method}, and the individual parts are described in the following section.

\subsection{Object Representation}
In the literature, several representations of shapes are used. The most common are voxel grids, point clouds, triangular meshes, or implicit surfaces. In our case, we decided to use a point cloud as input and to represent the object $O$ as an implicit surface expressed as
\begin{equation}
    \label{eq:object}
    O = \{ \mathbf{x} \in \mathbb{R}^3\; | \; f(\mathbf{x};\boldsymbol{\theta}, \mathbf{z}_r) = 0\},
\end{equation}
where $f: \mathbb{R}^3 \to \mathbb{R}$ represents a shape completion network and $\mathbf{z}_r$ is the latent vector optimized on points $\mathbf{x}$ from the current point cloud $\mathcal{X}_r$. Our approach is iterative (performs multiple grasps and shape completions), which is specified in the equations with the iteration number $r = 1,\dots,R$.

The completion network computes \acf{sdf}, i.e., the signed distance of each point to an underlying surface---details on the network and latent vectors are given in \secref{sec:completion_network}. Having a point cloud at the input allows for straightforward integration of visual and tactile point clouds, and the implicit surface at the output allows for extraction of complete shapes as a point cloud or a triangular mesh using the Marching Cubes algorithm~\citep{marching_cubes}. In the initial iteration, only visual information $\mathbf{v}_{init}$ from the RGB-D camera on a table is available. After each grasp attempt, new tactile data $\mathbf{t}_r$ and free space information $\mathbf{fs}_r$ are added to the point cloud. When a wrist RGB-D camera is available, visual information $\mathbf{v}_r$ is also added acquired from a pre-grasp view. After $r$ iterations, the aggregated point cloud used for shape completion is defined as $\mathcal{X}_r=\{\mathbf{v}_{init}, \mathbf{t}_1, \dots, \mathbf{t}_r, \mathbf{v}_1, \dots, \mathbf{v}_r\}$ together with the free space set $\mathbf{FS}_r = \{\mathbf{fs}_1, \dots, \mathbf{fs}_r\}$.

\subsection{Data Obtained After Grasp}
\label{sec:grasp_data}
Working with data obtained after grasp is an essential sub-problem of our method. After a successful grasp, the tactile information $\mathbf{t}_r$ can be obtained and added to the point cloud. When tactile sensors are in contact with the object, we compute their positions using forward kinematics and add their point cloud representations to the main point cloud $\mathcal{X}$---see \secref{sec:setup} for the details for each gripper we use. This information is often unavailable when the grasp attempt fails, although it can still be available if the gripper still closes around the object but the object subsequently slips out of the gripper during lifting.

\begin{figure}[htb]
    \centering
    \includegraphics[width=1\columnwidth]{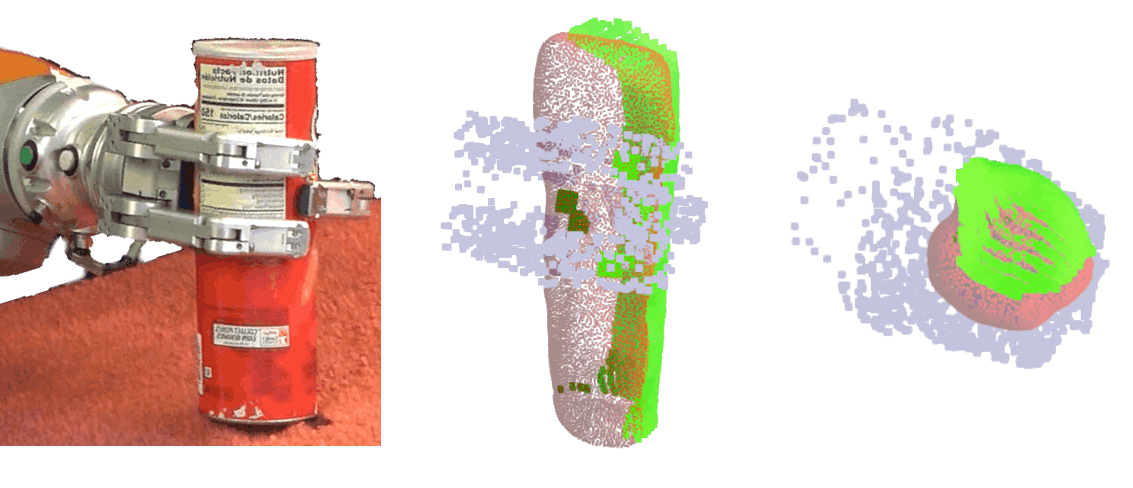}
    \caption{Example of real grasp (left) with front (middle) and top (right) view of the captured point cloud (green) with touches (green), reconstructed object (red) and collected free space (blue) for Barrett Hand.}
    \label{fig:free_space}
\end{figure}

The second form of information obtained is the so-called free space $\mathbf{fs}_r$, that is, the space that must be empty in the implicit representation of the object. \citet{Rustler2023VISHAC} used positions of the end-effector collected during the approach to perform an exploration action. Here, instead, we project a point cloud sampled from a gripper mesh to the actual pose of the gripper and use it as the free space---see \figref{fig:free_space} for reference. In other words, the space in the real world that is occupied by a gripper is translated into a space that must be free in the implicit representation of the object surface. This information is then used to constrain the neural network. These points must be outside the shape, i.e., have a positive signed distance. Thus, inside the network, the points can contribute to inference loss if their estimated signed distance is negative.

The last source of information is an optional wrist RGB-D camera. We can capture an additional RGB-D observation (in a pre-grasp pose; see the example in the left bottom part of \figref{fig:schema}), segment it, transform it into the reference
frame of $\mathcal{X}$, and fuse it with the other observations.

\subsection{Visuo-Haptic Shape Completion}
\label{sec:completion_network}
One of the core components of our method is the shape completion network. We build on the implicit surface approach from \citet{Rustler2023VISHAC}. The approach consists of a function $f(\mathbf{x};\boldsymbol{\theta}, \mathbf{z}_i):\mathbb{R}^3 \rightarrow \mathbb{R}$ that is a \ac{mlp} learned to approximate \ac{sdf}, with $\boldsymbol{\theta}$ being the parameters of the network and $\mathbf{z}_i$ a latent vector for a given shape $i \in I$. 

The function $f$ is represented by \ac{igr} from \citet{groppImplicitGeometricRegularization2020}. The method operates with a point cloud $\mathcal{X} = \{\textbf{x}_{1:C}\}$ and a set of normals $N = \{\mathbf{n}_{1:C}\}$, where, in both, $C$ is the number of points. During both training and test time, each shape $i\in I$ is encoded in a separate latent vector $\mathbf{z}_i$. Training is carried out with multiple shapes simultaneously, but inference is done with $|I| = 1$. The shapes (their latent vectors $\mathbf{z}_i$) are optimized iteratively in both the training and the test time. The trained weights $\boldsymbol{\theta}$ of the network are fixed during inference. The loss (for inference $|I| = 1$) is defined as
\begin{equation}
\label{eq:rec_loss}
\begin{split}
    \ell(\boldsymbol{\theta}, \mathbf{z}_i)
    &=\ell_\mathcal{X}(\boldsymbol{\theta}, \mathbf{z}_i)
    + \E_\mathbf{x}{\left[
    \norm{\nabla_\mathbf{x}f(\mathbf{x};\boldsymbol{\theta}, \mathbf{z}_i)}-1
    \right]}^2 \\
    &\quad + \norm{\mathbf{z}_i}
\end{split}
\end{equation}
where
\begin{equation}
\label{eq:X_loss}
\begin{split}
\ell_\mathcal{X}(\boldsymbol{\theta}, \mathbf{z}_i)
&=\frac{1}{C}\sum_{c=1}^C
\bigl(\abs{f(\mathbf{x}_c;\boldsymbol{\theta}, \mathbf{z}_i)} \\
&\quad + \norm{\nabla_\mathbf{x}f(\mathbf{x}_c;\boldsymbol{\theta}, \mathbf{z}_i)-\mathbf{n}_c}\bigr)
\end{split}
\end{equation}

The first term in \equationref{eq:rec_loss} encourages $f$ to vanish on $\mathcal{X}$ and $\nabla_{\mathbf{x}}f$ to be close to the supplied normals. The second term is called the Eikonal term and regularizes the network by pushing $\nabla_{\mathbf{x}}f$ to be of the unit Euclidean norm. More information about training and more specific features can be found in ~\citet{Rustler2023VISHAC,Rustler2022ActVH} and \citet{groppImplicitGeometricRegularization2020}.

\subsection{Segmentation and 6D Pose Estimation}
We aim to operate with generally unknown objects, which
makes segmentation and tracking non-trivial. For the purposes of our pipeline, we opted for the combination of GroundingDINO~\citep{liu2024grounding} for grounded object bounding box detection and SAM2~\citep{ravi2025sam} for RGB segmentation and tracking. This combination allows us to detect and track objects using a natural-language prompt (“an object on a table” in our experiments).

The input to the shape completion network is a segmented point cloud. We utilize the segmented 2D RGB image and deproject the corresponding pixels to 3D using depth obtained from a RealSense D435 camera. In our experience, this is faster and more robust than performing segmentation directly in 3D. Although our experiments use a single object at a time, the RGB-to-3D segmentation pipeline naturally extends to multiple objects and different environments by changing the language prompt.

After each grasp attempt, it is highly possible that the grasped object changes pose. To ensure that the new data are correctly aligned with the input point cloud $\mathcal{X}$, a 6D object pose is needed. Here, we only need the 6D pose with respect to the first one, i.e., to be able to detect pose changes with respect to the point cloud used for the completion of the first shape. Therefore, we obtain the segmented point cloud of the current scene and apply a point-to-point \ac{icp}~\citep{BeslICP} algorithm. Despite its simplicity, this approach was robust in our setting and does not require category- or instance-level mesh models. For abrupt pose changes (e.g., when an object falls and rolls), more advanced pose estimation may be beneficial.

\subsection{Grasp Generation}
The grasp generation algorithm is shown in \algoref{alg:grasp_generation}. The inputs are a triangular mesh of the object $M$ and the initial position of the given mesh. Random points on the surface of $M$ are uniformly sampled to obtain the point cloud $S$ with normals $N_S$. For each point $\mathbf{s}\in S$, a candidate gripper position $\mathbf{p}$ is obtained by shifting $\mathbf{s}$ along its normal by a set of predefined approach distances (\qtylist{10;11;12}{\centi\meter} in our experiments) corresponding to the distance from a virtual base frame of the gripper and the actual palm. The gripper orientation $\mathbf{o}$ is chosen so that the gripper's approach axis points opposite to the surface normal.

\begin{figure}[htb]
\begin{algorithm}[H]
\caption{Grasp Generation}
\label{alg:grasp_generation}
    \small
    \begin{algorithmic}[1] \Statex \textbf{Input:} Triangular mesh of the object $M$; initial position $\mathbf{l}$ of the object
    \Statex \textbf{Output:} Set of grasps $G$ with qualities
    \State $G = \emptyset$;
    \State $S, N_S$ = \textit{Sample($M$)}; \Comment{Obtain point cloud and normals}
    \For {$j = 0,\dots, |S|$}
        \For {$d \in D$} \Comment{Distances from base of the gripper}
        \label{alg:points_start}
            \State $\mathbf{p} = S[j] + d\cdot N_S[j]$; \Comment{Add pre-grasp offset}
            \State $\mathbf{o} = $ \textit{ComputeOrientation(}$-N_S[j]$\textit{)};
        \label{alg:points_end}
            \State \textit{LoadSimulation($M, \mathbf{l}, \mathbf{p}$)};
            \State \textit{TestGrasp($\mathbf{p}, \mathbf{o}$)};
            \State $q$ = \textit{ComputeQuality()}; \Comment{Based on \equationref{eq:quality}}
            \State $G = G \cup (\mathbf{p}, \mathbf{o}, q)$; \Comment{Add to final set}
        \EndFor
    \EndFor
    \State \textit{Sort($G$)}; \Comment{Sort by quality}
    \State \textbf{Return: } $G$
    \end{algorithmic}
\end{algorithm}
\end{figure}

The object $M$ is loaded into the simulation at a location $\mathbf{l}$ and the gripper at position $\mathbf{p}$ (with an additional offset of \qty{10}{\centi\meter} along the normal) with orientation $\mathbf{o}$. The gripper is then moved linearly towards the object to the original position $\mathbf{p}$. The gripper is closed, lifted upward and rotated $\pm 90^\circ$ around the gripper's approach axis.

For each grasp, we compute a grasp quality score
\begin{equation}
    \label{eq:quality}
    q = \frac{1}{|\Delta h_r| + ||\Delta \mathbf{l}|| + |\Delta h|} + \mathcal{F} + \mathcal{S} + \mathcal{T},
\end{equation}
where:
\begin{itemize}
    \item $|\Delta h_r|$ is an absolute value of the relative difference between the heights of the gripper and the object before and after lifting. This tests the slippage of the object in the gripper during lift up;
    \item $||\Delta \mathbf{l}||$ is the Euclidean distance between the positions of $M$ after and before rotation. This tests movement of object during rotation;
    \item $|\Delta h|$ is an absolute value of the difference in heights of $M$ after and before rotation. This gives a higher cost to vertical slippage during rotation;
    \item $\mathcal{S}$ is 1 when $M$ is still in contact with the gripper after rotation and 0 else;
    \item $\mathcal{F}$ is 1 when $M$ moved less than a predefined threshold;
    \item and optionally, $\mathcal{T}$ is the inverted absolute cosine between the gripper pose and the pose that would be for a top grasp (between 0 (top grasp) and 1 (side grasp)). We score side grasps better, as those are often more stable and gain more tactile information---this can be removed for general use.
\end{itemize}

The final set of grasps $G$ is sorted in descending order by qualities.

To speed up the process, multiple possible grasp poses are tested in parallel. Also, some tests can be interrupted without completion, e.g., when the gripper is in collision with the ground plane or the object from the beginning. Physics in our generator uses the \textit{PyBullet}~\citep{coumans2021} physics engine. Since grasps are evaluated in a gravity-enabled environment, the simulator supports parameters such as object mass and friction coefficients between the gripper and the object. We currently do not utilize those, and this remains for future work. We provide a code for our generator with the \textit{Robotiq 2F-85} and \textit{Barrett Hand} grippers.

\subsection{\methodname{} Algorithm}
\label{sec:method}

\begin{figure}[]
\begin{algorithm}[H]
\caption{\methodname{}}
\label{alg:pipeline}
    \begin{algorithmic}[1] \Statex \textbf{Input:} Number of iterations $R$
    \Statex \textbf{Output:} Shape $O$ of grasped object
    \State $v_{init} = CapturePointCloud()$; \label{line:SC_start}
    \State $\mathcal{X} = \{v_{init}\}$; \Comment{Visual-only point cloud}
    \State $F = \emptyset$; \Comment{Free space data}
    \State $O_0, M_0$ = \textit{CompleteShape($\mathcal{X}, F$)}; \label{line:SC_end}
    \State \textit{grasped = False} \Comment{Initialize end condition to False}
    \For {$r = 1:R$ }
        \State $\mathbf{T}$ = \textit{GetObjectPose()}; \Comment{ICP on current point cloud} \label{line:gr_start}
        \State $M_r$ = $inv(\mathbf{T})\cdot M_r$; \Comment{Transformation from canonical frame to current pose}
        \State $G_r$ = \textit{GenerateGrasps($M_r$)}; \Comment{Using \algoref{alg:grasp_generation}} \label{line:gr_end}
        \For {$\mathbf{g}, o \in G_r$} \Comment{$\mathbf{g}$ is a 6D pose of gripper with and o is a pre-grasp offset} \label{line:ac_gr_start}
            \If {\textit{IsFeasible($\mathbf{g}$)}}
                \State \textit{Move($\mathbf{g}$, o)}; \Comment{Move to pre-grasp pose}
                \If {\textit{UseWristCamera()}}
                    \State $\mathbf{v}_r$ = \textit{ExtractWristCameraData()};
                    \State $\mathcal{X} = \mathcal{X} \cup \mathbf{T}\cdot \mathbf{v}_r$;
                \EndIf
                \State \textit{Move($\mathbf{g}$)}; \Comment{Move to grasp pose}
                \State \textit{grasped = AttemptGrasp()}; \Comment{Close the jaws and check contacts} \label{line:ac_gr_end}
                \State $\mathbf{T}$ = \textit{GetObjectPose()};
                \State $F =F \cup \mathbf{T}\cdot\textit{ExtractFreeSpaceData()}$;\label{line:fs}
                \If {\textit{grasped}} \Comment{If jaws in contact with the object} \label{line:gr_eval_start}
                    \State $\mathbf{t}_r$ = \textit{ExtractTactileData()};
                    \State $\mathbf{T}$ = \textit{GetObjectPose()};
                    \State $\mathbf{t}_r$ = $\mathbf{T}\cdot \mathbf{t}_r$;
                    \State $\mathcal{X} = \mathcal{X} \cup \mathbf{t}_r$;
                    \State \textit{grasped = Lift()};
                \EndIf \label{line:gr_eval_end}
            \EndIf
        \EndFor
        \State $O_r, M_r$ = \textit{CompleteShape($\mathcal{X}, F$)};
        \If {\textit{grasped}} \Comment{End the pipeline}
            \State \textbf{break};
        \EndIf

    \EndFor
    \State \textbf{Return:} $O_r$
    \end{algorithmic}
\end{algorithm}
\end{figure}

The algorithm of our pipeline is provided in \algoref{alg:pipeline}. First, the initial visual information $\mathbf{v}_{init}$ is captured, the point cloud $\mathcal{X}$ used for shape completion is initialized with this information, and the complete point cloud $O_0$ and the triangular mesh $M_0$ are created (lines \ref{line:SC_start} to \ref{line:SC_end}). The triangular mesh $M_r$ is inserted into the grasp planner (the mesh is transformed from the canonical frame to the current pose in the real-world; lines \ref{line:gr_start}-\ref{line:gr_end}). The generated grasps are ordered by grasp quality, sequentially tested for feasibility, and the manipulator attempts the first feasible grasp. The robot first goes to pre-grasp pose (the same as grasp pose, but with offset along the normal). If a wrist RGB-D camera is available, an additional point cloud is captured at this pose. The robot then continues in the actual grasp pose and attempts a grasp (lines \ref{line:ac_gr_start}-\ref{line:ac_gr_end}).
The current pose $\mathbf{T}_r$ of the object is estimated and the free space data $\mathbf{fs}_r$ are collected (and transformed with the current pose $\mathbf{T}_r$ of the object; line \ref{line:fs}). If the grasp is successful (the object is in contact with the tactile sensors), the tactile data $\mathbf{t}_r$ are extracted, transformed with respect to the current pose of the object, and added to the overall point cloud $\mathcal{X}$.
The object is then lifted to test the stability (lines \ref{line:gr_eval_start}-\ref{line:gr_eval_end}). If the grasp is still ranked as successful, then the final shape is created and the pipeline stops. If the grasp is ranked unsuccessful in any of the two cases, the pipeline continues with the next iteration.

\subsection{Evaluation Metrics}
In the following metrics for shape completion evaluation, we use two point clouds $\mathbf{S}_1$ and $\mathbf{S}_2$ with their number of points $N, M$, respectively.

\textbf{\acf{cd}} is a standard metric to estimate the distance between two point clouds (shapes). The metric is defined as the average distance of each point in one set to the closest point in the second set and vice versa, i.e.,

\begin{align}
    d_{CD}(\mathbf{S}_1, \mathbf{S}_2) = \frac{1}{N}\sum_{\mathbf{x} \in \mathbf{S}_1}\min_{\mathbf{y} \in \mathbf{S}_2}\|\mathbf{x}-\mathbf{y}\|_2 + \nonumber \\ \frac{1}{M}\sum_{\mathbf{y} \in \mathbf{S}_2}\min_{\mathbf{x} \in \mathbf{S}_1}\|{\mathbf{x}-\mathbf{y}}\|_2,
\end{align}
where the lower the value, the better.

\textbf{\acf{hd}} is defined similarly to \ac{cd}, but now the output is the maximum distance (the worst case) between the closest points in two sets, i.e.,

\begin{align}
    d_{HD}(\mathbf{S}_1, \mathbf{S}_2) &= \max \left\{ \max_{\mathbf{x} \in \mathbf{S}_1} \min_{\mathbf{y} \in \mathbf{S}_2} \|\mathbf{x} - \mathbf{y}\|_2, \right. \nonumber \\
    &\qquad \qquad \left. \max_{\mathbf{y} \in \mathbf{S}_2} \min_{\mathbf{x} \in \mathbf{S}_1} \|\mathbf{x} - \mathbf{y}\|_2 \right\},
\end{align}

where the lower the value, the better.

\ac{cd} is more robust with respect to noise. However, as we deal with the comparison of two shapes without a single noisy point, \ac{hd} can show differences in shapes that could get lost when using the average in \ac{cd}---for example, missing handle of a larger cup.

\textbf{\acf{js}} is a standard metric for estimating the volumetric similarity of two shapes. It is defined as intersection over union of voxelized shapes, i.e.,
\begin{align}
    J(\mathbf{S}_1,\mathbf{S}_2) = \frac{|\mathbf{S}_1 \cap \mathbf{S}_2|}{|\mathbf{S}_1 \cup \mathbf{S}_2|},
\end{align}
where the values range from 0 to 1 and the higher the value, the better.

The \ac{cd} and \ac{hd} consider the boundaries of the shapes. Instead, \ac{js} takes into account the entire volume that the shapes occupy. In the general case, \ac{js} and \ac{cd} should correspond to each other, but there are cases where these do not agree. Imagine, for example, a long object with the longest axis parallel to the camera depth axis. If the shape completion has the correct overall shape but underestimates the depth, the \ac{cd} will be still lower, since the low error on the front face averages with the back error. On the other hand, \ac{js} will be quite low, as the shared volume is low.

\textbf{Precision, Recall, and $\mathbf{F}_\mathbf{1}$ score} are typically used in classification. Here, we use the definition from NVIDIA Kaolin Library~\citep{KaolinLibrary}: a point is counted as a \ac{tp} if it is within a radius $r$ of a point in the other set. \ac{fn} are \gt{} points farther than $r$ from the completion, and \ac{fp} are completion points farther than $r$ from the \gt{}. And then
\begin{equation}
    F_1 = \frac{2TP}{2TP + FP + FN} = 2\cdot\frac{precision \cdot recall}{precision + recall},
\end{equation}
where the values range from 0 to 1 and the higher the value, the better.

Intuitively, precision reflects how much additional noise (e.g., hallucinated peaks, added handle) is added to the completion and recall says how many points/features are missing (e.g., missing handle). Together, the $F_1$ score assesses reconstruction quality by penalizing both hallucinated and missing geometry.

\textbf{\acf{gsr}} is a simple metric for assessing grasping. It is computed as the ratio of successful grasps and attempted grasps. \section{Experiments and Results}
\label{sec:exps}
The experiments demonstrate real-world grasping and visuo-haptic shape completion in a closed-loop setting. We compare our pipeline against baselines in terms of both reconstruction quality and grasp success. Example runs are shown in the accompanying video.
We evaluated the method on a set of 9 objects from the \textit{YCB Dataset}~\citep{Calli2015} with distinct shapes, sizes, and materials.  During the experiments, paper boxes and plastic containers were filled with pasta to gain higher rigidity to reduce penetration during grasping, both to protect the objects from damage and to make grasping comparable to rigid objects. The transparent spray bottle was filled with water to keep the transparency. We performed 10 runs of the entire pipeline for each object. We performed all our experiments on a workstation with NVIDIA RTX 4080 Super and Intel i7-14700K.

\subsection{Setup}
\label{sec:setup}
The experiments were carried out on two robots:
\begin{enumerate}[label=(\roman*)]
    \item 7-\ac{dof} Kinova Gen3 robotic manipulator with a two-finger Robotiq 2F-85 gripper equipped with a Contactile PapillArray tactile sensor. Each finger contains a $3\times3$ grid of pillars measuring 3D contact forces. For each activated pillar (force above a fixed threshold), we add a \qtyproduct{5 x 5}{\milli\meter} square patch to the tactile point cloud.
    \item 7-\ac{dof} KUKA LBR iiwa 7 R800 robotic manipulator with three-finger Barrett Hand equipped with capacitive tactile sensors. The hand contains 96 taxels (24 on each finger and on the palm). For each activated taxel, we add a rectangular patch of size \qtyproduct{10.5 x 10.5}{\milli\meter}, \qtyproduct{5 x 5}{\milli\meter}, or \qtyproduct{5 x 3}{\milli\meter} depending on whether it is a palm, finger, or fingertip sensor.
\end{enumerate}

\begin{figure}[htb]
    \centering
    \includegraphics[width=0.975\columnwidth]{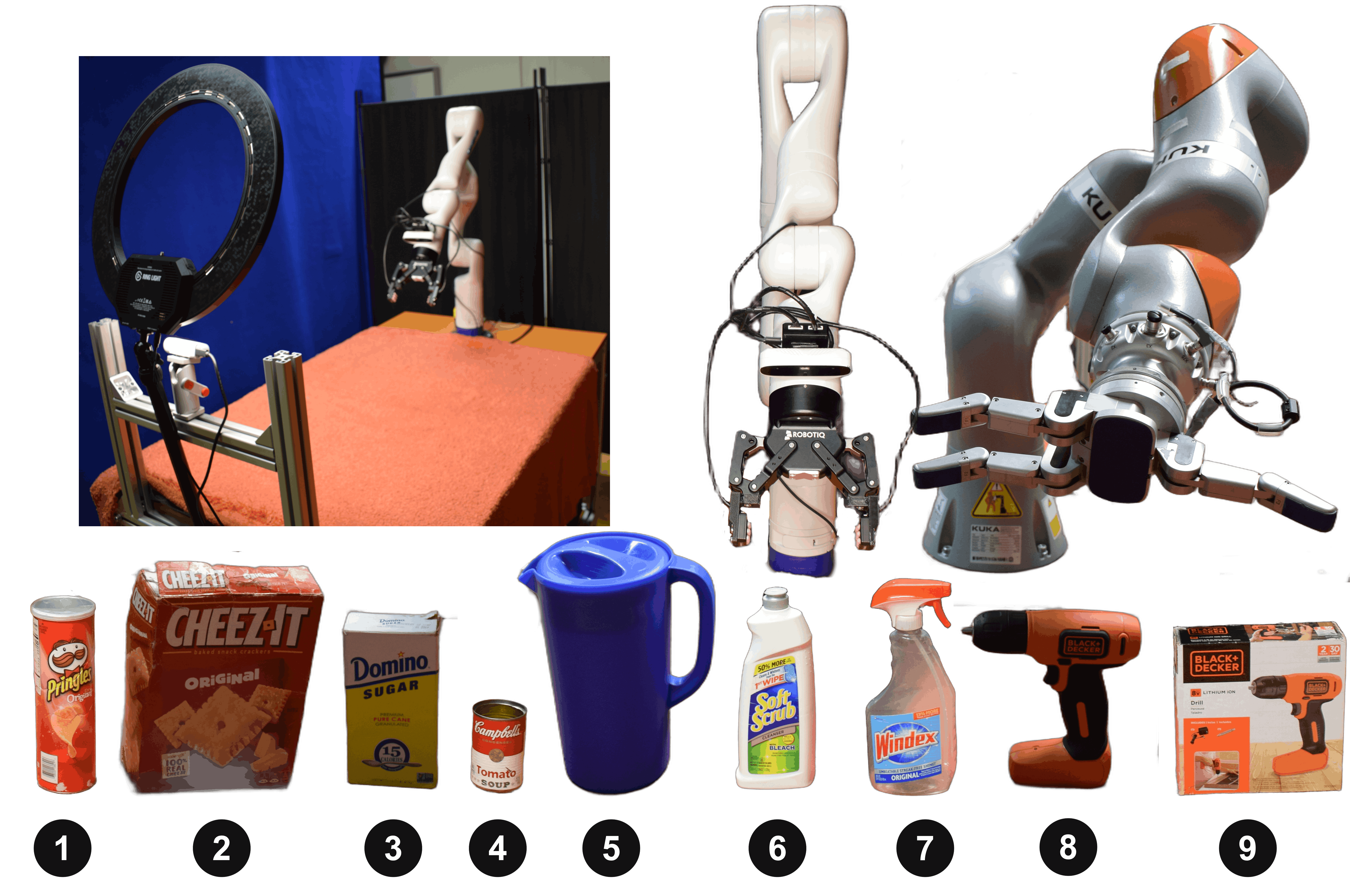}
    \caption{The table with robot, light and camera (top left); the robots used (top right)---Kinova Gen3 on the left and KUKA LBR iiwa 7 R800 on the right---and the objects used in the experiments (bottom), .}
    \label{fig:real_setup}
\end{figure}

In addition, the Intel RealSense D435 RGB-D camera was placed on the table in front of the objects (the objects were always positioned \qty{60}{\centi\meter} from the camera) and the robot. In addition, we utilized artificial light behind the camera to maintain a constant illuminance of \qty{600}{\lux} near the objects. The experimental setup, robot, camera, light, and all objects used can be seen in \figref{fig:real_setup}.

\subsection{Shape Completion}
\label{sec:shape_completion}
We first evaluate the shape completion capabilities of \methodname{}. The final result is characterized by both shape completion and active perception, and thus it is not evaluation only of the shape completion network, but of the whole pipeline. For both grippers, we provide per object comparison with ablation study on the information used---visual-only, visual + free space, visual + tactile, and visual + tactile + free space. The following plots and tables show data for runs in which grasp was marked as successful in any iteration of the pipeline. To compute point-based metrics, the reconstruction must be aligned to a \gt{} point cloud. We first estimate the alignment between the \gt{} scan and the initial camera point cloud using \ac{icp}~\citep{BeslICP}. A small manual correction was then applied to compensate for residual calibration bias; the resulting transform was reused for all completions (different iterations or different information used) starting from the same initial camera view.

\begin{table*}[h]
\centering
\caption{Shape completion metrics. The data are averaged from averages of the 9 objects used, i.e., the standard deviation shows deviation per object. Averaged performances for each object were computed from runs where grasp was marked successful in any iteration. The best values for each column are marked \textbf{bold} and the second best are \underline{underlined}---rows with gray background are excluded for fair comparison. N/A stands for not available values and dash (-) for values not computed because of completed shape format.}
\label{tab:shape_completion}
\resizebox{\textwidth}{!}{

\begin{tabular}{l cccc c cccc}
\toprule
& \multicolumn{4}{c}{\textbf{Barrett Hand}} & & \multicolumn{4}{c}{\textbf{Robotiq 2F-85}} \\
\cmidrule(lr){2-5} \cmidrule(lr){7-10}

\textbf{Method} &
\textbf{JS} [\%] $\uparrow$ & \textbf{CD} [mm] $\downarrow$ & \textbf{F$_1$} [\%] $\uparrow$ & \textbf{HD} [mm] $\downarrow$ & &
\textbf{JS} [\%] $\uparrow$ & \textbf{CD} [mm] $\downarrow$ & \textbf{F$_1$} [\%] $\uparrow$ & \textbf{HD} [mm] $\downarrow$ \\
\midrule

\textbf{\methodname{}} & & & & & & & & & \\
\hspace{1em}\textit{Visual-Only} & \underline{$55.84 \pm 7.25$} & \underline{$19.89 \pm 3.70$} & $63.45 \pm 5.75$ & \underline{$39.73 \pm 6.94$} & & $54.67 \pm 6.84$ & $20.26 \pm 3.01$ & $63.19 \pm 5.05$ & $38.60 \pm 6.37$ \\
\hspace{1em}\textit{Free Space} & $52.44 \pm 6.64$ & $21.31 \pm 3.20$ & $61.31 \pm 4.83$ & $42.69 \pm 7.40$ & & $55.32 \pm 7.91$ & $19.69 \pm 2.77$ & $63.73 \pm 4.36$ & \underline{$38.08 \pm 7.33$} \\
\hspace{1em}\textit{Tactile} & $\mathbf{59.54 \pm 7.28}$ & $\mathbf{18.25 \pm 3.87}$ & $\mathbf{66.50 \pm 8.55}$ & $\mathbf{38.91 \pm 8.05}$ & & $\mathbf{58.14 \pm 6.61}$ & $\mathbf{18.64 \pm 2.21}$ & \underline{$65.02 \pm 4.67$} & $\mathbf{36.53 \pm 5.74}$ \\
\hspace{1em}\textit{Visuo-Haptic} & $53.56 \pm 6.92$ & $20.75 \pm 3.30$ & $61.91 \pm 5.23$ & $39.98 \pm 4.46$ & & \underline{$55.64 \pm 7.67$} & \underline{$19.58 \pm 2.25$} & $63.79 \pm 3.74$ & $38.15 \pm 6.70$ \\

\rowcolor{gray!10} \hspace{1em}\textit{Wrist} & N/A & N/A & N/A & N/A & & $63.89 \pm 8.49$ & $16.23 \pm 3.01$ & $71.20 \pm 7.03$ & $32.14 \pm 7.69$ \\
\rowcolor{gray!10} \hspace{1em}\textit{Wrist Tactile} & N/A & N/A & N/A & N/A & & $63.00 \pm 8.85$ & $16.52 \pm 3.15$ & $70.16 \pm 7.37$ & $32.61 \pm 7.50$ \\
\addlinespace

\textbf{VISHAC} \citeyearpar{Rustler2023VISHAC} & & & & & & & & & \\
\rowcolor{gray!10} \hspace{1em}Touch 5 & N/A & N/A & N/A & N/A & & $60.07 \pm 4.50$ & $18.65 \pm 3.64$ & $63.92 \pm 9.22$ & $37.68 \pm 13.58$ \\
\rowcolor{gray!10} \hspace{1em}Touch 6 & N/A & N/A & N/A & N/A & & $60.75 \pm 4.82$ & $18.35 \pm 3.84$ & $65.03 \pm 9.75$ & $37.46 \pm 12.59$ \\
\rowcolor{gray!10} \hspace{1em}Touch 15 & N/A & N/A & N/A & N/A & & $66.75 \pm 7.13$ & $15.51 \pm 3.29$ & $72.09 \pm 10.09$ & $33.23 \pm 8.47$ \\

\textbf{HyperPCR} \citeyearpar{Rosasco2022} & & & & & & & & & \\
\hspace{1em}\textit{Visual-Only} & -- & $26.03 \pm 6.00$ & \underline{$66.07 \pm 8.60$} & $60.44 \pm 17.93$ & & -- & $24.60 \pm 5.02$ & $\mathbf{70.08 \pm 8.78}$ & $60.66 \pm 16.15$ \\
\hspace{1em}\textit{Tactile} & -- & $23.85 \pm 6.53$ & $63.55 \pm 9.76$ & $55.31 \pm 16.87$ & & -- & $24.67 \pm 4.84$ & $62.26 \pm 6.49$ & $58.20 \pm 18.96$ \\
\addlinespace

\textbf{3DSGrasp} \citeyearpar{mohammadi_3dsgrasp_2023} & & & & & & & & & \\
\hspace{1em}\textit{Visual-Only} & -- & $40.96 \pm 8.88$ & $41.34 \pm 8.02$ & $68.52 \pm 15.33$ & & -- & $39.77 \pm 7.15$ & $42.69 \pm 9.41$ & $68.76 \pm 14.42$ \\
\hspace{1em}\textit{Tactile} & -- & $37.65 \pm 7.40$ & $42.48 \pm 7.16$ & $66.22 \pm 14.37$ & & -- & $36.90 \pm 7.37$ & $44.06 \pm 8.88$ & $66.62 \pm 15.01$ \\
\addlinespace

\textbf{VQDIF} \citeyearpar{yan2022ShapeFormerTransformerBasedShapea} & & & & & & & & & \\
\hspace{1em}\textit{Visual-Only} & $17.07 \pm 6.10$ & $54.02 \pm 5.08$ & $19.40 \pm 3.13$ & $78.69 \pm 8.07$ & & $32.31 \pm 6.37$ & $40.59 \pm 4.28$ & $25.11 \pm 4.19$ & $57.35 \pm 9.94$ \\
\hspace{1em}\textit{Tactile} & $19.51 \pm 6.85$ & $52.64 \pm 4.29$ & $19.30 \pm 1.71$ & $79.84 \pm 7.67$ & & $32.60 \pm 6.34$ & $40.59 \pm 4.30$ & $25.15 \pm 4.37$ & $57.51 \pm 9.71$ \\

\bottomrule
\end{tabular}}
\end{table*}

\begin{figure*}[h]
    \centering
    \includegraphics[width=1\textwidth]{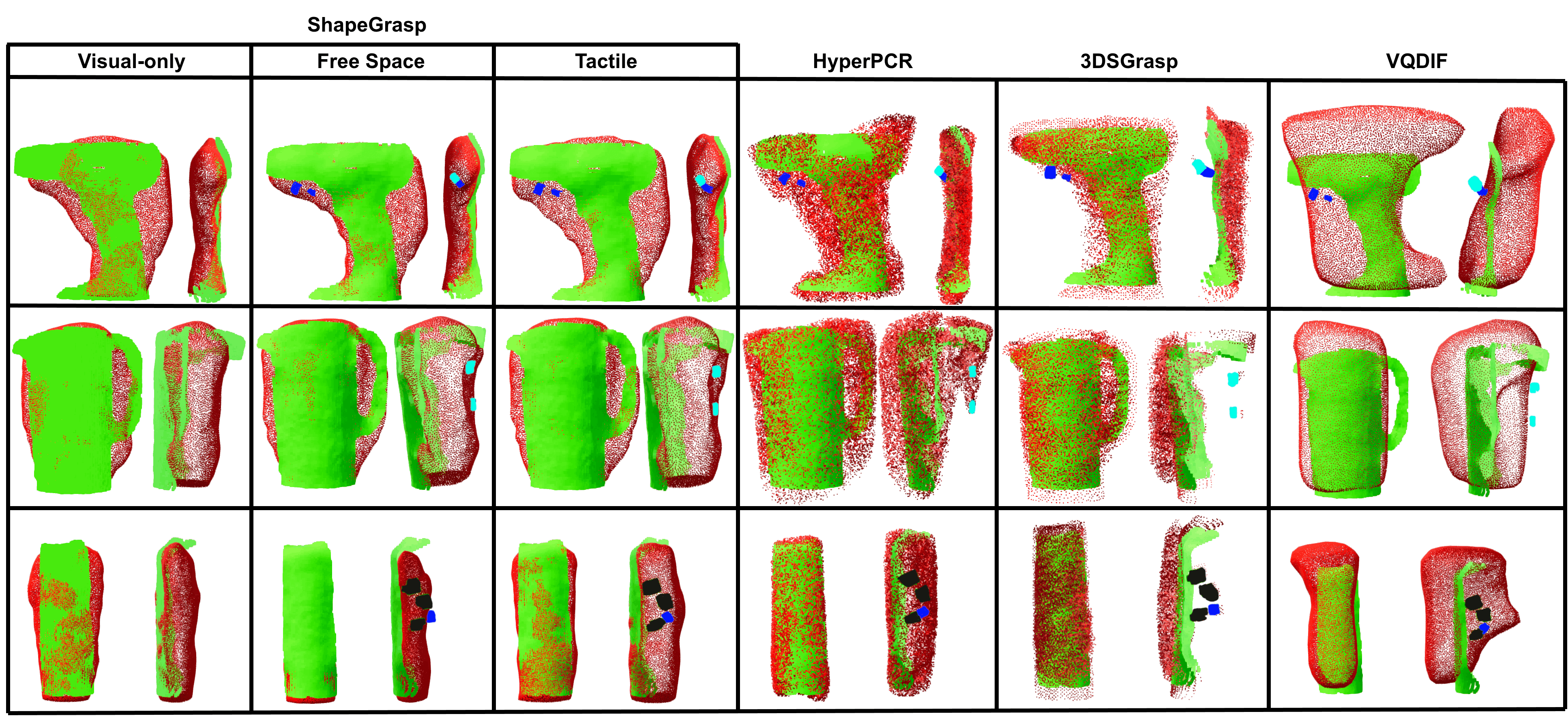}
    \caption{Examples of front (left) and side (right) views of visual point clouds (green) with touches (highlighted in black if normals are facing towards current point-of-view, blue if facing opposite, and cyan if facing side-ways; see the accompanying video for 3D visualization) and reconstruction (red) for \textit{Visual-Only}, \textit{Free Space} and \textit{Tactile} versions of \methodname{} and the baselines.}
    \label{fig:completions}
\end{figure*}

We compare against three shape completion networks: HyperPCR~\citep{Rosasco2022}, 3DSGrasp~\citep{mohammadi_3dsgrasp_2023} and VQDIF~\citep{yan2022ShapeFormerTransformerBasedShapea}. For these baselines, we use the visual and tactile point clouds collected during our experiments. We additionally compare with VISHAC~\citep{Rustler2023VISHAC}, a visuo-haptic completion pipeline that performs dedicated haptic exploration; for VISHAC we evaluate the meshes and point clouds released by the authors\footnote{\url{https://osf.io/j6rkd}}.

In the comparison, we compare several versions of our method based on the information used for shape completion:
\begin{enumerate*}[label=(\roman*)]
    \item \textit{Visual-Only} -- only point cloud from table camera is used;
    \item \textit{Free Space} -- point cloud from table camera and free space (the space that is occupied by the gripper during grasp that thus must be empty in the object implicit representation);
    \item \textit{Tactile} -- point cloud from the table camera and tactile information;
    \item \textit{Visuo-Haptic} -- point cloud from table camera, tactile and free space information;
    \item \textit{Wrist} -- point cloud from the table camera and point cloud from the wrist camera (applicable only for Kinova robot with Robotiq 2F-85 gripper);
    \item \textit{Wrist Tactile} -- point cloud from both cameras and tactile information.
\end{enumerate*}

\subsubsection{Overall comparison}
\tabref{tab:shape_completion} shows the comparison, where the best values in columns are shown in bold and the second best underlined---we did not include the \textit{wrist} versions of \methodname{} and VISHAC in this comparison, as it would not be fair as both acquire much more new information compared to our method. We also do not report \ac{js} for HyperPCR and 3DSGrasp as our \ac{js} implementation requires watertight shapes, which are not provided by the two methods. Thus, the \ac{js} performance for the methods would be very poor and would not be a fair comparison.

We can see that \methodname{} (specifically the \textit{Tactile} version) is the best for both grippers in almost every metric, except for the $F_1$ score. VQDIF and 3DSgrasp are far from our performance. In the case of VQDIF, it is probably caused by the fact that all the objects are outside of the train set (which is, however, the fact also for us) and expect bigger chunks of objects---our point clouds are quit thin (in depth). For 3DSGrasp, some of the objects are from the training set and some are outside the set. Examples are shown in \figref{fig:completions}. We can see that 3DSGrasp mostly scatters points near the input point cloud and even usually to the \enquote{outside} of the objects.

HyperPCR is closer to our results, mainly in \ac{cd} and $F_1$ score. But it is more than \qty{2}{\centi\meter} worse in \ac{hd}. All of these can be seen in the examples in \figref{fig:completions}. Firstly, HyperPCR creates completed shapes tighter to the input point cloud than our method, which helps with \ac{cd} and makes $F_1$ higher as the precision is high---see the drill in \figref{fig:completions} where our completions are very \enquote{wide}. On the other hand, it scatters some points \enquote{outside} of the input point cloud, making \ac{cd} higher than ours. For some objects---see chip can in \figref{fig:completions}---it correctly completes the depth. But in some cases---see pitcher in \figref{fig:completions}---there is a large missing volume, which makes \ac{hd} high.

\begin{figure}[htb]
    \centering
    \includegraphics[width=0.9\columnwidth]{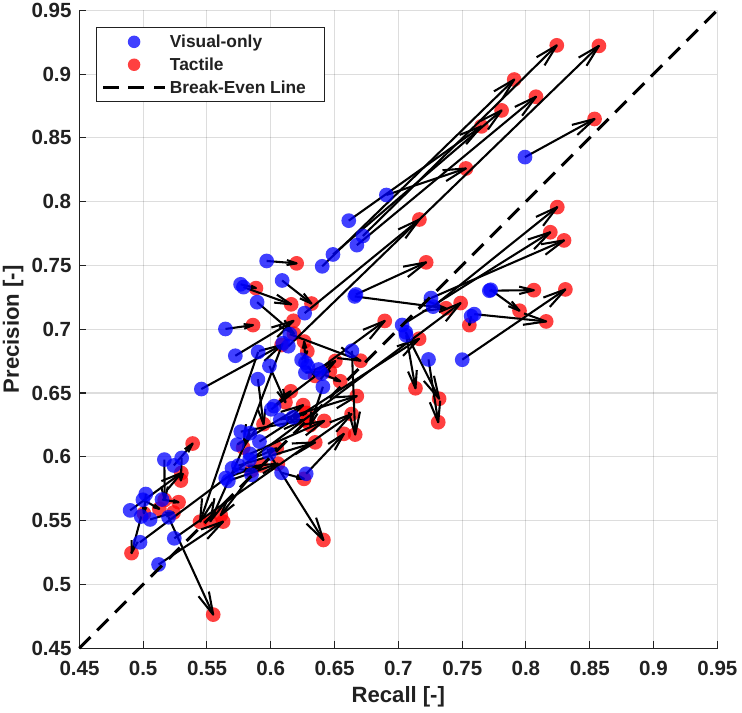}
    \includegraphics[width=0.9\columnwidth]{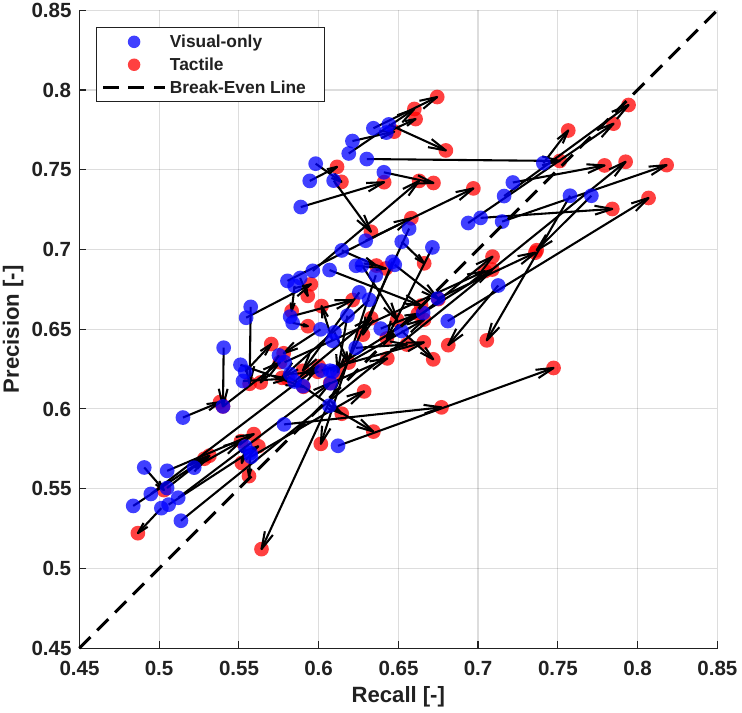}
    \caption{Precision and Recall for Barrett Hand (top) and Robotiq 2F-85 (bottom) for all objects. Only runs where grasp was marked as successful in any iteration are shown. The diagonal line shows the optimal precision/recall ratio.}
    \label{fig:precrec}
\end{figure}

In general, we can say that \methodname{} is better in estimating the total depth of objects and does not create points outside the input point cloud boundary. However, it inflates the shapes to the sides, which mainly influences the $F_1$ score. It is visible in \figref{fig:precrec} where we show the evolution of precision and recall for the \textit{Tactile} version of \methodname{}. We can see that visual-only reconstruction (blue points) is for both grippers more on the precision side of the plot, i.e., it is missing some points from the \gt{}, but those guessed are more often correct. After adding tactile information (red points), either both precision and recall increases (which should be ideal case) or recall increases, but precision decreases, i.e., the method guesses more points (more depth in our case) but also adds more points that should not be there (the inflation).

We need to note that HyperPCR wins in terms of speed. It takes only about \qty{0.2}{\second} for it to output completion and our method requires about \qty{5}{\second}, which is a notable difference. HyperPCR only outputs \enquote{scattered} point clouds, while \methodname{} outputs watertight meshes that are needed for the grasp planner.

\begin{figure*}[htb]
    \centering
    \includegraphics[width=1.4\columnwidth]{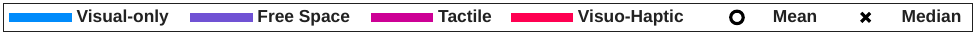}\\
    \includegraphics[width=1\columnwidth]{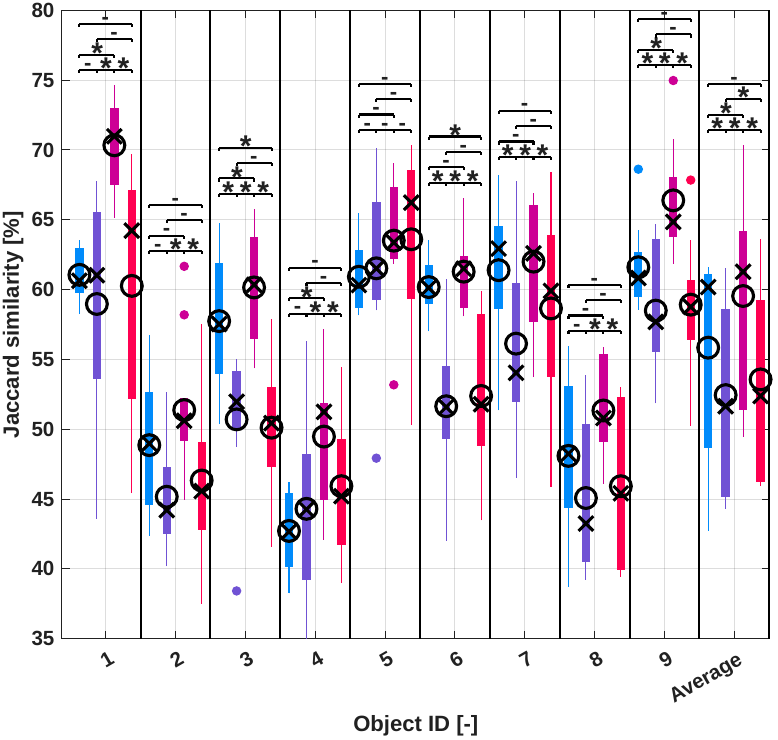}
    \includegraphics[width=1\columnwidth]{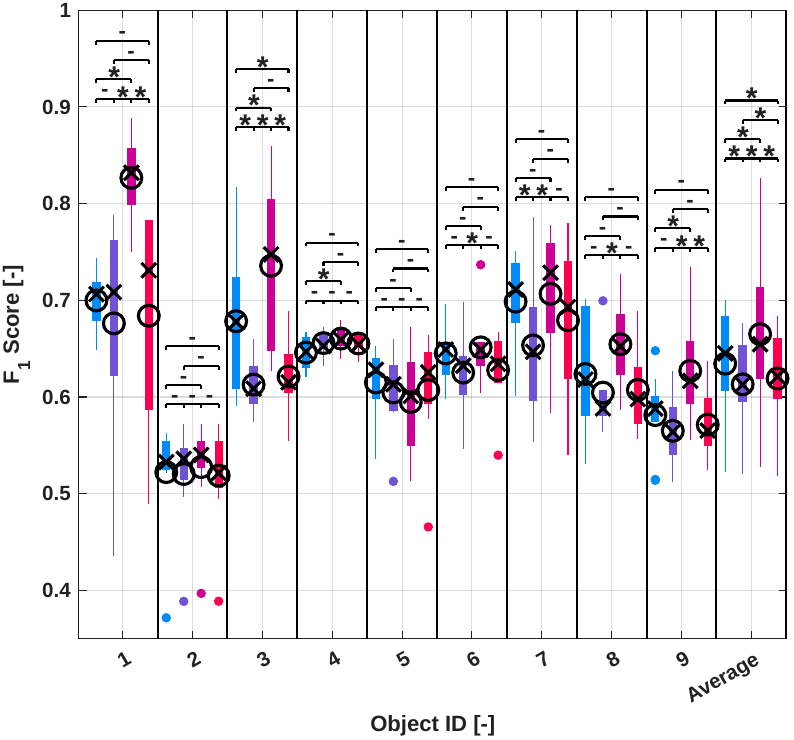}
    \includegraphics[width=1\columnwidth]{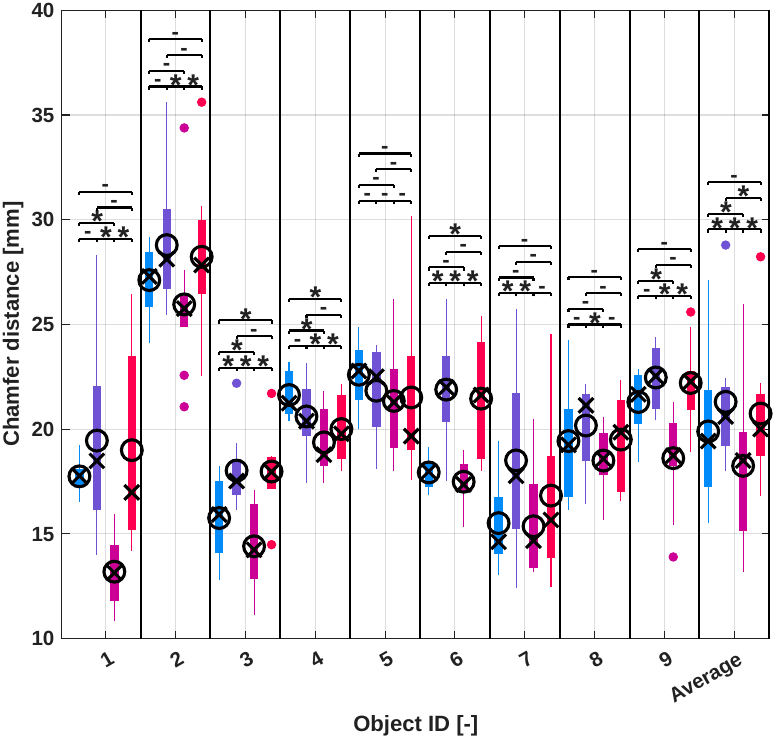}
    \includegraphics[width=1\columnwidth]{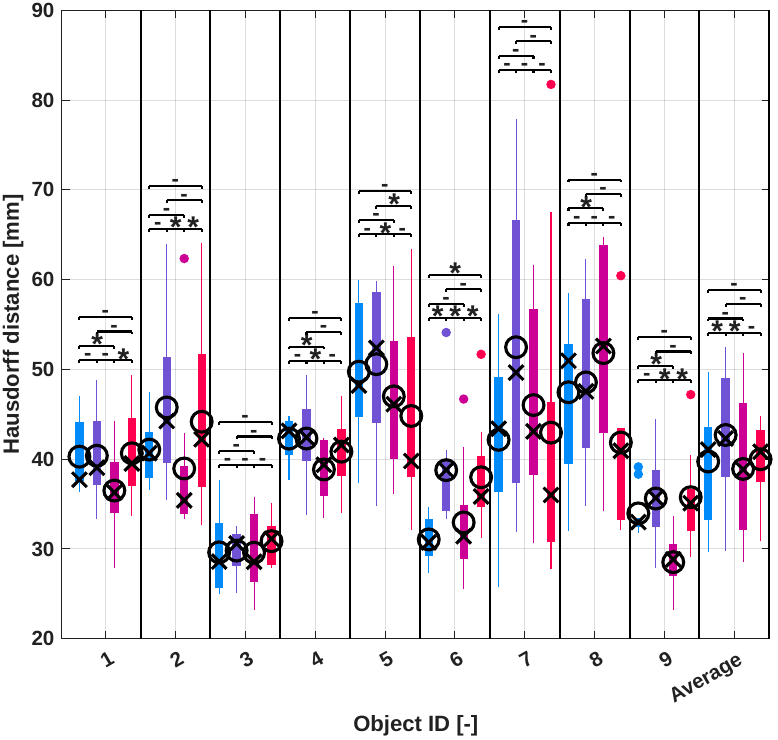}
    \caption{Shape completion metrics for Barrett Hand. The boxplots show the 25th to 75th percentile in bold, non outlier values with whiskers and outliers with points. Pair-wise differences were evaluated using the Wilcoxon signed-rank test---asterisks (*) denote significant differences ($p < 0.05$), while dashes (–) indicate non-significant results. For Jaccard similarity (top left) and $F_1$ score (top right) the higher the value, the better. For Chamfer distance (bottom left) and Hausdorff distance (bottom right) the lower the value, the better. Object IDs correspond to IDs in \figref{fig:real_setup}. Data for each object are computed from runs where the grasp was successful in any iteration. The average data are computed from averages of the individual objects.}
    \label{fig:kuka_sc}
\end{figure*}

\begin{figure*}[htb]
    \centering
\includegraphics[width=1.4\columnwidth]{imgs/legend_sc.pdf}\\
    \includegraphics[width=1\columnwidth]{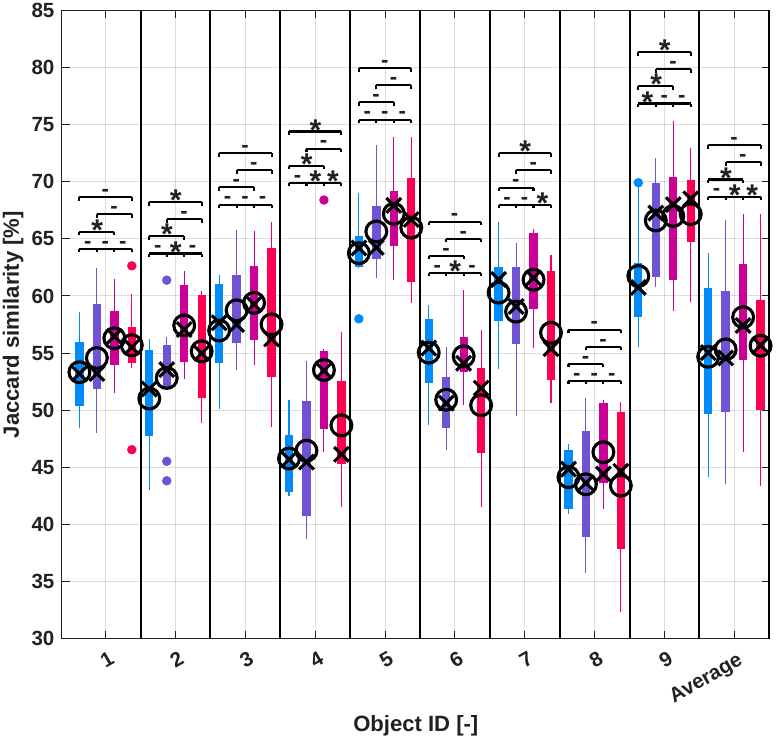}
    \includegraphics[width=1\columnwidth]{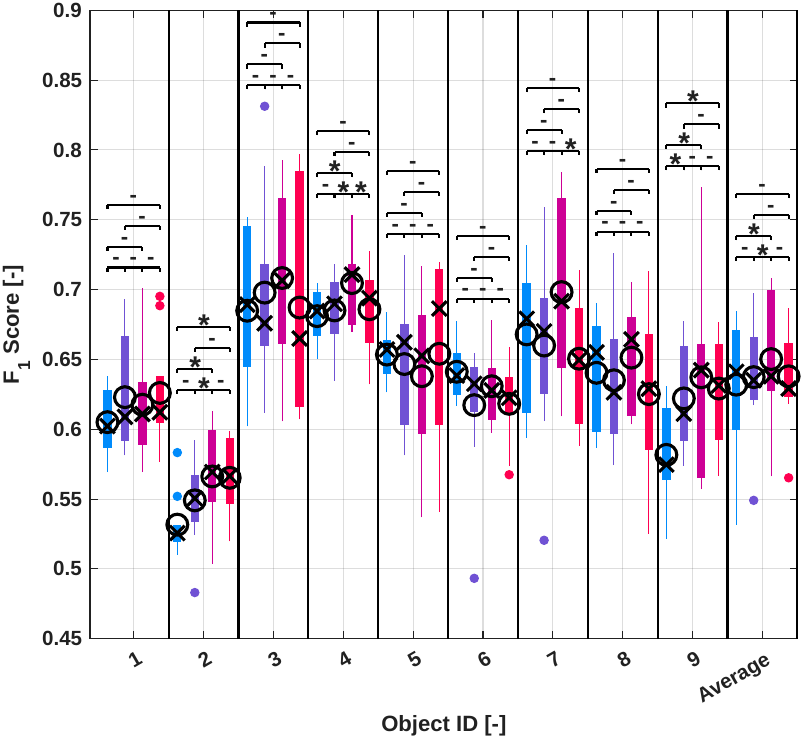}
    \includegraphics[width=1\columnwidth]{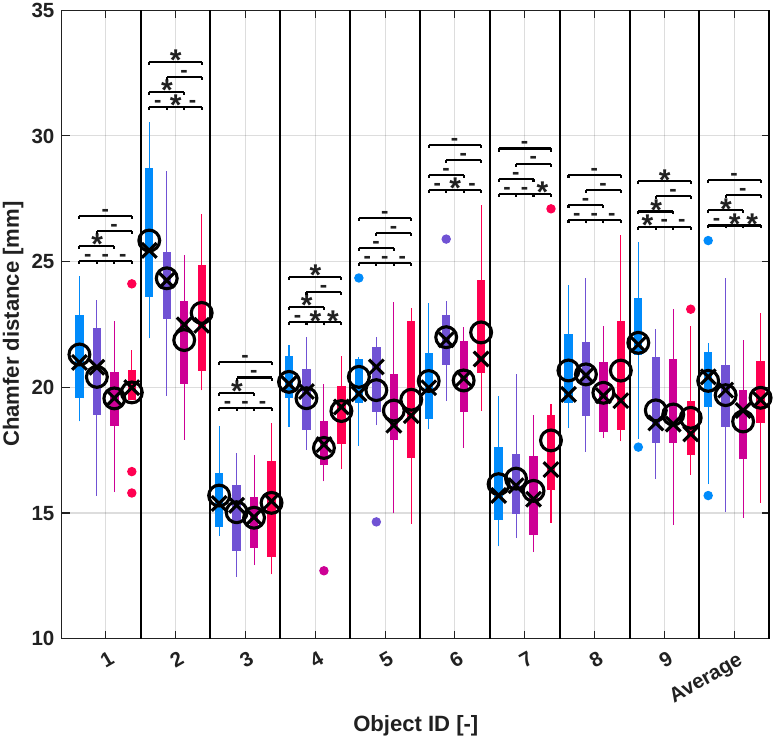}
    \includegraphics[width=1\columnwidth]{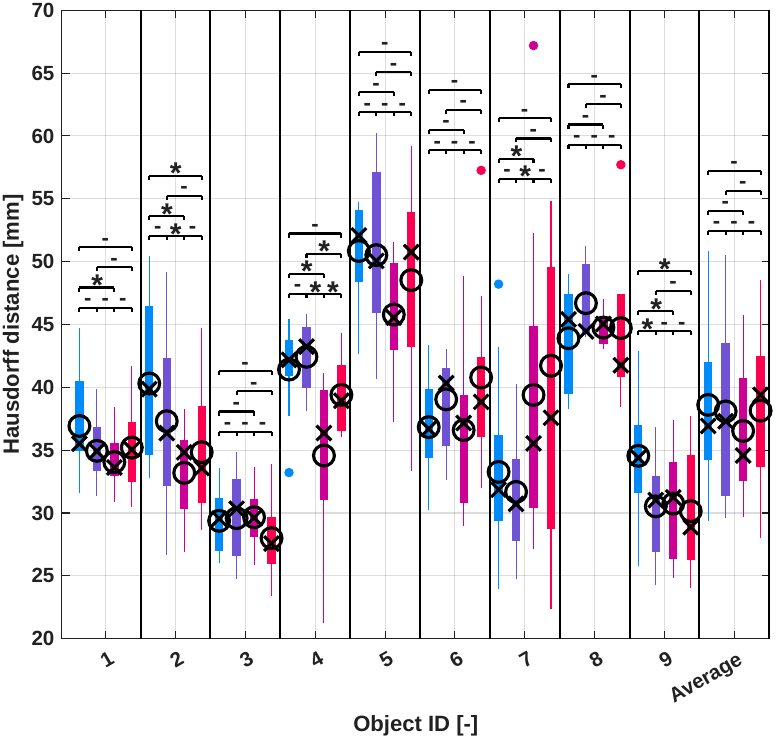}
    \caption{Shape completion metrics for Robotiq 2F-85 gripper. The boxplots show the 25th to 75th percentile in bold, non outlier values with whiskers and outliers with points. Pair-wise differences were evaluated using the Wilcoxon signed-rank test---asterisks (*) denote significant differences ($p < 0.05$), while dashes (–) indicate non-significant results. For Jaccard similarity (top left) and $F_1$ score (top right) the higher the value, the better. For Chamfer distance (bottom left) and Hausdorff distance (bottom right) the lower the value, the better. Object IDs correspond to IDs in \figref{fig:real_setup}. Data for each object are computed from runs where the grasp was successful in any iteration. The average data are computed from averages of the individual objects.}
    \label{fig:kinova_sc}
\end{figure*}

\subsubsection{Wrist camera}
The Kinova Gen3 robot was used to gather a second visual viewpoint before grasp execution. Depth sensing requires the camera to be sufficiently far from the object (the nominal minimum range is around \qty{50}{\centi\meter}, although it works even closer). We captured wrist RGB-D data from a pre-grasp pose before moving to the final grasp pose---with the offsets of \qtylist{20;15;10}{\centi\meter} based on feasibility. When the only feasible pre-grasp pose was \qty{10}{\centi\meter} from the grasp pose, depth was often unreliable; therefore, wrist point clouds were available only in 50 of 90 runs. As shown in \tabref{tab:shape_completion}, adding wrist-camera observations improves reconstruction metrics by roughly 10\% compared to the \textit{Tactile} variant---initial point cloud from the table camera and tactile information from the gripper. This is expected because additional visual views provide more global information without requiring contact. Nevertheless, wrist RGB-D data are not always available (e.g., no wrist camera, adverse lighting, or specular/transparent objects), in which case grasp-derived tactile information provides a useful alternative.

\subsubsection{The effect of the information collected during grasping} \tabref{tab:shape_completion} and \figref{fig:kuka_sc}--\figref{fig:kinova_sc} summarize the ablation study across information sources. The plots also provide pair-wise statistical significance between the results. We use Wilcoxon signed rank test as we have only 10 repetitions for each and we cannot assume normal distribution.

\textbf{Free Space.} Our initial assumption was that free space will have a lower positive effect than tactile information, but the combination will be superior. Interestingly, the results suggest that free space could even affect the results negatively. In \tabref{tab:shape_completion} we can see that for Robotiq 2F-85 gripper free space information helps slightly with quality (e.g., an increase from 54.7\% to 55.3\% in \ac{js}), but if we look at \figref{fig:kinova_sc}, the difference is statistically significant ($p<0.05$) only for object ID 9 for all metrics and is not statistically significant in average performance for any metric. For Barrett Hand, it actually decreases performance (e.g., decrease of \ac{js} to 52.4\% from 55.8\%) and from \figref{fig:completions} we can see that it is more often statistically significant (and is significant when averaged over all objects).

\figref{fig:completions} gives some insight into the problem. We can see that for the drill, the free space actually helped the reconstruction as the drill is very stiff and therefore the fingers of the gripper stopped \enquote{outside} of the object and carved some of the inflated volume near the handle. The other extreme appears with the chips can (last object in \figref{fig:completions}), where the gripper crushed the object too much and ended \enquote{inside}, making the reconstruction extremely bad. The examples in \figref{fig:completions} are captured with the Barrett hand, which is more powerful than the other gripper, and, more importantly, it can clasp around the object (as in the case of the chips can). This difference from the 2F-85 gripper (which has two relatively small fingers that can add only limited amount of free space) is, in our opinion, the reason Barrett Hand free space actually decreased the performance. Still, we believe that free space is promising information (mainly if a given gripper lacks the tactile information), but the objects must be either really firm or the grasp must be precise and not penetrate into objects.

\textbf{Tactile.} If we focus on the \textit{Tactile} version (initial point cloud and tactile data from grasp), \figref{fig:kuka_sc} and \figref{fig:kinova_sc} reveal that the difference from the visual-only method is statistically significant for 4 out of 9 objects for the \ac{js}, \ac{cd}, and $F_1$ scores. However, the average difference is significant for all three metrics. The situation is different for \ac{hd}, where the situation for individual objects is similar, but the average difference is not statistically significant. This is a reasonable result, as \ac{hd} is created from one worst point of the completed shape. Thus, it is highly unlikely that local tactile information improves that specific point (or cluster of points, in general). If we now focus on the actual improvements, the average improvement for both grippers and all metrics is 5.5\%. That may not seem a lot at first sight, but absolute performance corresponds to five to six touches of VISHAC---see the data in \tabref{tab:shape_completion}. VISHAC utilizes a similar shape completion network and also starts from a single-view point cloud. Then it touches the object in the most uncertain place. And touching the object five times takes almost 3 minutes.

\textbf{Visuo-Haptic.} The combination of initial point cloud, tactile data, and free space provides results better than \textit{Free Space} and worse than \textit{Tactile}. Again, for Robotiq it performs better than visual-only version and for Barrett it performs worse. This proves the fact that the free space provides, in the current setup, a more negative impact than tactile data.

\subsection{Grasping}
Here, we evaluate the grasping performance. We compare the methods (ours and the baselines) as unified pipelines of shape completion network and grasp planner. Thus, for example, \methodname{} in the results represents the whole pipeline with one or more shape completions, and not only the grasp planner.

In our experiments, we performed 10 full runs of our pipeline (initial shape completion with one or more grasp attempts and shape completions with new data) for each object, resulting in 90 runs per robot/gripper. The grasps were marked as successful if the object remains in the gripper after being lifted \qty{10}{\centi\meter} from the surface of the table. We report two success rates: \methodnameo{} (success after the first grasp attempt) and \methodnamen{} (success after the final attempt, after up to $R$ iterations). The number of attempts per run therefore ranges from 1 to $R$. In some cases, the motion planner marks an approach pose (\qty{10}{\centi\meter} from the grasp pose) feasible, but no collision-free Cartesian path to the final grasp pose is found; such attempts are counted as unsuccessful for that iteration.

We compare our approach with several \sota{} baselines that also deal with shape completion and test their reconstruction with their or \sota{} grasp planners. We further divided the comparison according to the gripper used: Barrett Hand or Robotiq 2F-85.

\subsubsection{Barrett Hand}

\begin{table}[h]
\centering
\caption{Grasp success rates for Barrett Hand. The object IDs corresponds to IDs in \figref{fig:real_setup}. The best result per line is shown in \textbf{bold} and the second best \underline{underlined}. The mean values are computed over the number of attempts and the average row is computed over the individual objects. Dash (-) stands for object not being in the dataset.}
\label{tab:barrett_grasping}
\fontsize{7.5pt}{8pt}\selectfont
\setlength{\tabcolsep}{0.75pt}
\renewcommand{\arraystretch}{1.1}
\begin{tabular}{@{}lcccc@{}}
\toprule
\textbf{Object ID} & \begin{tabular}[c]{@{}c@{}}Varley\\(\citeyear{varleyShapeCompletionEnabled2017})\\from\\\citeyearpar{lundellRobustGraspPlanning2019}\end{tabular} & \begin{tabular}[c]{@{}c@{}}USN\\\citeyearpar{lundellRobustGraspPlanning2019}\end{tabular} & \methodnameo{} & \methodnamen{} \\ \midrule
\textbf{1} & 0.5 & \underline{0.8} & \underline{0.8} & \textbf{0.9} \\
\textbf{2} & - & - & 0.8 & \textbf{1} \\
\textbf{3} & \textbf{1} & 0.9 & 0.7 & 0.8 \\
\textbf{4} & - & - & 0.5 & \textbf{0.7} \\
\textbf{5} & 0.2 & 0.4 & \underline{0.7} & \textbf{0.8} \\
\textbf{6} & 0.6 & \underline{0.7} & 0.5 & \textbf{0.9} \\
\textbf{7} & 0.7 & \underline{0.8} & 0.5 & \textbf{0.9} \\
\textbf{8} & 0.1 & \underline{0.4} & \textbf{0.6} & \textbf{0.6} \\
\textbf{9} & \underline{0.7} & 0.5 & \underline{0.7} & \textbf{1} \\
\midrule
\textbf{Average} & 0.54 & \underline{0.64} & \underline{0.64} & \textbf{0.84} \\ \midrule
\textbf{Attempts} & 100 & 100 & 10 & 10 \\ \midrule
\textbf{Time [s]} & 16.53 & 83.25 & 7.05$\pm$1.80 & 9.25$\pm$5.00 \\ \midrule
\begin{tabular}[c]{@{}l@{}}\textbf{Grasp} \\ \textbf{planner}\end{tabular} & \begin{tabular}[c]{@{}c@{}}GraspIt!\\\citeyearpar{miller2004GraspIt}\end{tabular} & \begin{tabular}[c]{@{}c@{}}GraspIt!\\\citeyearpar{miller2004GraspIt}\end{tabular} & \methodname{} & \methodname{} \\ \bottomrule
\end{tabular}
\end{table}

The first gripper to compare is the Barrett Hand. We compare our performance with USN~\citep{lundellRobustGraspPlanning2019} and \citet{varleyShapeCompletionEnabled2017}---the grasps for both were planned with \graspit{}~\citep{miller2004GraspIt} and we took the results from~\citep{lundellRobustGraspPlanning2019}. The results are shown in \tabref{tab:barrett_grasping}. We can compare on 7 objects used in common. We can see that \methodnamen{} is the best in 6 out of 7 cases. The only case where the baselines perform better is object number 3---smaller paper box. Our success rate for that object is 80\% and the two unsuccessful cases were caused by falling of the object during grasp---our pipeline was able to estimate the object pose after the fall and propose new grasps, but those were marked as infeasible due to safety limits. For \methodnameo{}, the method USN performs on average the same---the difference is visible in the comparison per-object.

Our method requires substantially shorter grasp planning time. \methodnameo{} requires on average \qty{7.05 \pm 1.80}{\second} (computed over the first grasp attempt in all 90 runs), while the baseline requires \qty{83.25}{\second}. For \methodnamen{} the average time is \qty{9.25\pm5.00}{\second} (computed from 76 runs with a successful grasp in any iteration) as it was necessary to plan multiple times in some runs. For the cases where the first grasp was unsuccessful, $2.33 \pm 0.57$ grasps were required for a successful one, resulting in the average of $1.31\pm 0.63$ grasp on the 76 successful runs of the pipeline.

\begin{table*}[htb]
\centering
\caption{Grasp success rates for Robotiq 2F-85 gripper. The object IDs corresponds to IDs in \figref{fig:real_setup}. The best result per row is shown in \textbf{bold} and the second best \underline{underlined}. The mean values are computed over the number of attempts and the average row is computed over the individual objects. N/A stands for non-provided value and dash (-) for object not being in the dataset. For Act-VH and VISHAC, the first averaged line represent results with purely visual shape completion and the second row results after performing several haptic exploration actions.}
\label{tab:robotiq_grasping}
\renewcommand{\arraystretch}{1.1}
\resizebox{\textwidth}{!}{\begin{tabular}{lccccccccc}
\toprule
\textbf{Object ID} & 
\begin{tabular}[t]{@{}c@{}}USN~\citeyearpar{lundellRobustGraspPlanning2019}\\from \citeyearpar{Rosasco2022}\end{tabular} & 
\begin{tabular}[t]{@{}c@{}}HyperPCR\\\citeyearpar{Rosasco2022}\end{tabular} & 
\begin{tabular}[t]{@{}c@{}}3DSGrasp\\\citeyearpar{mohammadi_3dsgrasp_2023}\end{tabular} & 
\begin{tabular}[t]{@{}c@{}}Duarte\\\citeyearpar{duarte_measuring_2025}\end{tabular} & 
\begin{tabular}[t]{@{}c@{}}Duarte\\\citeyearpar{duarte_measuring_2025}\end{tabular} & 
\begin{tabular}[t]{@{}c@{}}Act-VH\\\citeyearpar{Rustler2022ActVH}\end{tabular} & 
\begin{tabular}[t]{@{}c@{}}VISHAC\\\citeyearpar{Rustler2023VISHAC}\end{tabular} & 
\methodnameo{} & 
\methodnamen{} \\ \midrule
\textbf{1} & 0.5 & \textbf{1} & \underline{0.8} & 0.66 & 0.52 & 0.33 & N/A & \textbf{1} & \textbf{1} \\
\textbf{2} & - & - & - & - & - & \textbf{1} & N/A & \textbf{1} & \textbf{1} \\
\textbf{3} & 0.9 & \textbf{1} & - & - & - & \textbf{1} & N/A & 0.80 & \textbf{1} \\
\textbf{4} & - & - & - & - & - & \textbf{1} & N/A & 0.90 & \textbf{1} \\
\textbf{5} & 0.6 & \textbf{0.9} & 0.4 & 0.28 & 0.4 & 0.66 & N/A & 0.50 & \underline{0.7} \\
\textbf{6} & 0.6 & \underline{0.9} & 0.7 & 0.72 & 0.3 & \textbf{1} & N/A & 0.80 & \textbf{1} \\
\textbf{7} & 0.6 & 0.7 & - & - & - & 0.66 & N/A & 0.50 & \textbf{1} \\
\textbf{8} & 0.7 & \textbf{1} & \underline{0.8} & 0.64 & 0.63 & 0.66 & N/A & 0.30 & 0.5 \\
\textbf{9} & 0.8 & 0.8 & 0.8 & 0.52 & 0.33 & 0.66 & N/A & \underline{0.90} & \textbf{1} \\ \midrule
\textbf{Average} & 0.67 & \underline{0.90} & 0.70 & 0.56 & 0.44 & \begin{tabular}[c]{@{}c@{}}0.38\\0.77\end{tabular} & \begin{tabular}[c]{@{}c@{}}0.63\\0.85\end{tabular} & 0.74 & \textbf{0.91} \\ \midrule
\textbf{Attempts} & 10 & 10 & 10 & 50 & 50 & 3 & 3 & 10 & 10 \\ \midrule
\textbf{Time [s]} & N/A & N/A & 6 & 2 & 2 & N/A & N/A & 5.63$\pm$0.86 & 7.33$\pm$4.01 \\ \midrule
\textbf{\begin{tabular}[t]{@{}l@{}}Grasp\\Planner\end{tabular}} & 
\begin{tabular}[t]{@{}c@{}}GPD\\\citeyearpar{tenpas2017GraspPoseDetection}\end{tabular} & 
\begin{tabular}[t]{@{}c@{}}GPD\\\citeyearpar{tenpas2017GraspPoseDetection}\end{tabular} & 
\begin{tabular}[t]{@{}c@{}}GPD\\\citeyearpar{tenpas2017GraspPoseDetection}\end{tabular} & 
\begin{tabular}[t]{@{}c@{}}GPD\\\citeyearpar{tenpas2017GraspPoseDetection}\end{tabular} & 
\begin{tabular}[t]{@{}c@{}}PointNetGPD\\\citeyearpar{liang2019PointNetGPDDetectingGrasp}\end{tabular} & 
\begin{tabular}[t]{@{}c@{}}GraspIt!\\\citeyearpar{miller2004GraspIt}\end{tabular} & 
\begin{tabular}[t]{@{}c@{}}GraspIt!\\\citeyearpar{miller2004GraspIt}\end{tabular} & 
\methodname{} & \methodname{} \\ \bottomrule
\end{tabular}}
\end{table*}

\subsubsection{Robotiq 2F-85}
In the Robotiq 2F-85 experiments, we compare against methods that use as the grasp planner:
\begin{enumerate}[label=(\roman*)]
    \item \graspit{}~\citep{miller2004GraspIt} -- Act-VH and VISHAC by \citet{Rustler2022ActVH,Rustler2023VISHAC}
    \item GPD~\citep{tenpas2017GraspPoseDetection} or PointNetGPD~\citep{liang2019PointNetGPDDetectingGrasp} -- HyperPCR by \citet{Rosasco2022}, 3DSGrasp by \citet{mohammadi_3dsgrasp_2023}, work of \citet{duarte_measuring_2025}, and USN by \citet{lundellRobustGraspPlanning2019}. 
\end{enumerate}
The results are summarized in \tabref{tab:robotiq_grasping}. The baseline results are taken from the corresponding publications except for USN for which the results are taken from the work of \citet{Rosasco2022}. 

First, we can compare with methods that use \graspit{} to generate possible grasps. The baseline methods---Act-VH and VISHAC---are visuo-haptic shape completion networks and provide two types of results:
\begin{enumerate*}[label=(\roman*)]
    \item visual-only completion (shown in the first row of averaged results) that is comparable to \methodnameo{}. As the completion networks in the baselines are very similar to ours, this basically compares the pure performance of our grasp planner and \graspit{};
    \item visuo-haptic completion (shown in the second row of averaged results) that used shape after several haptic actions that is comparable to \methodnamen{}. This compares the overall success rates of entire pipelines---both shape completion and grasp planning.
\end{enumerate*}
We observe that \methodnameo{} performs better than the visual-only version of both baselines. In particular, since we build on a modified version of the VISHAC completion network, this suggests that the proposed grasp planner can exploit the reconstructed geometry more effectively than \graspit{}. The visuo-haptic results of Act-VH and VISHAC achieve higher success than \methodnameo{} (77\% and 85\% vs. 74\%), but require approximately \qty{300}{\second} for the required haptic exploration. In contrast, \methodnamen{} achieves the highest mean success (91\%) while operating in a grasp-attempt loop.

If we look at the other baselines, even \methodnameo{} performs better than most of them except for HyperPCR~\citep{Rosasco2022}. The baselines utilized GPD~\citep{tenpas2017GraspPoseDetection} or PointNetGPD~\citep{liang2019PointNetGPDDetectingGrasp}, which are \sota{} point cloud-based grasp generation pipelines, and grasp generation is faster than our method. \methodnameo{} requires \qty{5.63\pm0.86}{\second} to generate grasp proposals and the baselines need under \qty{2}{\second}---the first two baselines do not comment on the time needed, but GPD itself usually requires less than a second; the authors of 3DSGrasp~\citep{mohammadi_3dsgrasp_2023} state that their pipeline takes about \qty{6}{\second} for grasp planning, but we assume that it also contains motion planning. However, neither GPD nor PointNetGPD allows one to specify additional details, such as weight or friction, that are important in real-world grasping.

When comparing \methodnamen{} with HyperPCR, the \ac{gsr} results are basically equal. HyperPCR achieves 90\% \ac{gsr} and \methodnamen{} 91\% when averaged over all 9 objects, or 88\% when averaged over 7 objects that are common between the methods. The main discrepancy in the object-wise comparison is for the object 8---power drill. We achieve only 50\% \ac{gsr} compared to 100\% of HyperPCR. This object is very stiff, heavy (more than \qty{600}{\gram}) and prone to fall, and therefore the grasp pose must be very precise. In our experiments, the object fell down in all the unsuccessful cases---in some of them the pose estimation algorithm failed; in other trials, all grasps were infeasible and thus subsequent grasp attempts did not help. We believe that the poor performance is caused mainly by insufficient completion quality for this object (see \figref{fig:completions}) where we can see that completion for the drill is inflated to the sides, which made the grasp planner often grasp only the edge of the handle making the object fall. In terms of grasp proposal speed, our method required \qty{7.33\pm4.01}{\second} for the 82 successful grasps---plus the time spent on potential unsuccessful grasp when more than one grasp was needed. 
 \section{Conclusion, Discussion, and Future Work}
We presented \methodname{}, a visuo-haptic shape completion pipeline tightly connected to a physics-based grasp planner. 
A grasp proposal is created based on initial, visual-only shape completion. Regardless of the success of the grasp, new information is collected, and the reconstructed shape is adjusted. In case of unsuccessful grasp, our proposed pipeline is able to detect that, estimate new pose of the object, and attempt a new grasp using the refined shape. 

We showed that this symbiotic relationship beats baselines in terms of both shape completion quality and grasp success rate. We tested the pipeline in the real world using two robots and two different tactile sensorized grippers---two-finger Robotiq 2F-85 gripper and three-fingered Barrett Hand. We used 9 objects from the YCB set~\citep{Calli2015} that were selected to span different sizes, materials, and shapes. We evaluated the method both quantitatively and qualitatively using four shape completion metrics and one grasping metric. In shape completion, compared to the best baseline method, HyperPCR~\citep{Rosasco2022}, our method achieves over 30\% lower error in some metrics, e.g., \qty{18.25}{\milli\meter} \acf{cd} for our method and \qty{23.85}{\milli\meter} for the baseline. The combination of multiple metrics allowed us to properly understand the strength and weaknesses of all the methods. \methodname{} achieves better performance by respecting the initial point cloud more, but tends to overestimate the sides of reconstructed objects. 

We further provided an ablation study on the influence of different information that can be added to guide the completion---tactile information from gripper, free space resulting from the morphology and pose of the gripper, and, optionally, visual information from a wrist camera. Unsurprisingly, additional visual information from a wrist camera offers superior performance. However, using tactile information provides an average improvement of 5.5\% (averaged over all four metrics, tested objects, and both robots/grippers) with statistically significant (p-value using Wilcoxon signed rank test $<0.05$) difference from visual-only completion. 

In terms of grasping, we achieved the best grasp success rate for both grippers, with 84\% success for three-fingered Barrett Hand and 91\% for two-fingered Robotiq gripper. This shows the versatility of the grasp proposal pipeline. And, importantly, the baselines used \sota{} grasping pipelines such as GPD~\citep{tenpas2017GraspPoseDetection}, PointNetGPD~\citep{liang2019PointNetGPDDetectingGrasp}, or \graspit{}~\citep{miller2004GraspIt} and we still outperformed those.

Although the overall results are promising and outperform \sotan{}, there are still shortcomings. As discussed in the results section, after inputting tactile information into the network, it sometimes tends to prioritize recall over precision, making the final shapes bloated, which influences both completion quality and grasping. In addition, our grasp planner supports the use of information about physical properties such as friction or weight, but it is not used at the moment. An initial guess of the properties can be inferred using vision and then improved given the physical contact (see \citet{Kruzliak2024}), which could improve both the first and subsequent grasp attempts. The greatest weakness of \methodname{} is the processing time. Both shape completion and grasp planning are time demanding---around \qty{5}{\second} each (see \secref{sec:exps} for details on the hardware used). This could be improved for better real-time performance, mainly by replacing some computationally heavy parts of grasp planning with approaches based on \ac{ml}. Finally, our pipeline is designed to be useful in downstream tasks that require better knowledge about the grasped object (such as robot-human handover task) that is hard to be obtained without a proper object model. However, testing \methodname{} in such scenarios remains to be explored later.

Both code and data are available online at \url{https://rustlluk.github.io/ShapeGrasp}. 
\bmhead{Statements and Declarations}

\textbf{Funding.} This work was co-funded by the European Union under the project Robotics and Advanced Industrial Production (reg. no. CZ.02.01.01/00/22\_008/0004590). L.R. was additionally supported by the Czech Science Foundation (GAČR) under project No. 26-22606S and Grant Agency of the Czech Technical University in Prague, grant No. SGS26/075/OHK3/1T/13.

\textbf{Competing interests.} The authors have no competing interests to declare that are relevant to the content of this article.

\textbf{Author contributions.} Lukas Rustler and Matej Hoffmann contributed to the study conception and design. Lukas Rustler implemented the method, performed the experiments, analyzed the data, and wrote the first draft of the manuscript. Matej Hoffmann supervised the work and reviewed and edited the manuscript. Both authors read and approved the final manuscript.

\textbf{Data and code availability.} Data and code are available at \url{https://rustlluk.github.io/ShapeGrasp}.

\bibliography{refs}
\end{document}